\newif\iffollowingorders
\followingordersfalse
\newif\ifcomments
\commentstrue

\iffollowingorders
\documentclass{article}

\usepackage[final]{nips_2018}

\usepackage[utf8]{inputenc} 
\usepackage[T1]{fontenc}    
\usepackage{hyperref}       
\usepackage{url}            
\usepackage{booktabs}       
\usepackage{amsfonts}       
\usepackage{nicefrac}       
\usepackage{microtype}      
\else
\documentclass[letterpaper,11pt]{article}
\usepackage{fullpage}
\usepackage[utf8]{inputenc} 
\usepackage{times}
\usepackage[T1]{fontenc}
\usepackage{bm}             
\usepackage{amsfonts}       
\fi

\usepackage{xfrac}

\usepackage{paralist} 
\usepackage{psfrag}
\usepackage{wrapfig}
\usepackage{pstricks} 
\iffollowingorders
\psset{unit=0.38cm}
\else
\psset{unit=0.47cm}
\fi
\psset{linewidth=0.05}%
\psset{fillstyle=solid}%

\def\mypapertitle{Adversarial Risk and Robustness: General Definitions \\ and  Implications for the Uniform Distribution\footnote{This is the full version of a work with same title that will appear in NIPS 2018.}}

\iffollowingorders
\def\mypaperauthors{Dimitrios I. Diochnos\thanks{Authors have contributed equally.}\\
University of Virginia\\
\texttt{diochnos@virginia.edu}
\And Saeed Mahloujifar\footnotemark[\value{footnote}]\\
University of Virginia\\
\texttt{saeed@virginia.edu}
\And Mohammad Mahmoody\thanks{Supported by NSF CAREER award CCF-1350939 and University of Virginia's SEAS Research Innovation Awards.}\\
University of Virginia\\
\texttt{mohammad@virginia.edu}}

\else
\def\mypaperauthors{Dimitrios I. Diochnos\thanks{Authors have contributed equally.}\\
University of Virginia\\
\texttt{\small diochnos@virginia.edu}
\and Saeed Mahloujifar\footnotemark[\value{footnote}]\\
University of Virginia\\
\texttt{\small saeed@virginia.edu}

\and Mohammad Mahmoody\thanks{Supported by NSF CAREER award CCF-1350939 and University of Virginia SEAS Research Innovation Award.}\\
University of Virginia\\
\texttt{\small mohammad@virginia.edu}}

\date{}
\fi

\iffollowingorders
\usepackage{xcolor}
\else
\usepackage{paralist}
\usepackage[numbers,square,sort&compress]{natbib}

\usepackage[
   colorlinks%
   ,plainpages=false
   ,hypertexnames=true
   ,naturalnames
   ,hyperindex
   ,linkcolor=blue
   ,citecolor=blue
   ,urlcolor=RoyalBlue
   ,urlcolor=blue
   ,pdfauthor={}
   ,pdftitle={}
   ,pdfsubject={Adversarial Perturbations}
]{hyperref}
\fi

\usepackage[
  english 
  , ruled 
  , linesnumbered 
  , noline 
  , noend 
]{algorithm2e}
\usepackage{upgreek}

\definecolor{RoyalBlue}{cmyk}{1, 0.50, 0, 0}
\definecolor{ForestGreen}{cmyk}{0.864, 0.0, 0.429, 0.396}
\definecolor{Brown}{cmyk}{0.0,0.692,0.925,0.529}

\newcommand{\Mnote}[1]{\ifcomments{\color{red} [\bf {M:}  #1]}\fi}

\usepackage{amssymb,amsmath,amsthm}
\usepackage[makeroom]{cancel}
\usepackage{thmtools,thm-restate}
\usepackage{comment}
\usepackage{multirow}
\usepackage{float}

\usepackage{centernot}

\newcommand{\PS}{\mathsf{BSize}}
\newcommand{\ERegion}{\mathcal{E}}

\newcommand{\erf}{\mathsf{erf}}

\newcommand{\vol}{\mathsf{vol}}
\newcommand{\intB}{\mathcal{IB}}
\newcommand{\extB}{\mathcal{EB}}
\newcommand{\metric}{\mathbf{d}}
\newcommand{\Ball}{\mathcal{B}all}

\newcommand{\PC}{^{\mathrm{PC}}}
\newcommand{\CI}{^{\mathrm{CI}}}
\newcommand{\ER}{^{\mathrm{ER}}}

\newcommand{\HD}{\mathsf{HD}}

\newcommand{\Rob}{\mathsf{Rob}}

\newcommand{\eDR}{^{\mathrm{ER}}}

\newcommand{\nomPC}{^{\mathrm{PC}}_{r}}
\newcommand{\nomCI}{^{\mathrm{CI}}_{r}}
\newcommand{\nomDR}{^{\mathrm{ER}}_{r}}

\newcommand{\nomtPC}{\PC}
\newcommand{\nomtCI}{\CI}
\newcommand{\nomtDR}{\ER}

\newcommand{\C}{\cC}

\newcommand{\problem}{\ensuremath{\mathsf{P}}\xspace}

\newcommand{\Rplus}{\mathbf{R}_+}
\newcommand{\Risk}{\mathsf{Risk}}

\newcommand{\fhat}[2]{\ifthenelse{\equal{#2}{}}{\hat{f}[#1]}{\ifthenelse{\equal{#2}{0}}{\hat{f}[\emptyset]}{\hat{f}[#1_{\leq #2}]}}}
\newcommand{\gain}[2]{\ifthenelse{\equal{#2}{}}{g[#1]}{g[#1_{\leq #2}]}}

\newcommand{\parag}[1]{{\bf #1}}

\newcommand{\pr}[2][]{\Pr_{\ifthenelse{\isempty{#1}}{}{{#1}}}\left[{#2}\right]}

\makeatother

\newcommand{\sm}{\setminus}

\newcommand{\e}{{e}}

\usepackage{graphicx,accents}

\newcommand{\remove}[1]{}

\newcommand{\se}{\subseteq}

\newcommand{\floor}[1]{\lfloor #1 \rfloor}

\newcommand{\set}[1]{\{ #1 \}}

\newcommand{\bits}{\{0,1\}}

\newcommand{\size}[1]{\left|#1\right|}
\newcommand{\abs}[1]{\size{#1}}

\newcommand{\To}{\mapsto}

\newcommand{\R}{{\mathbb R}}
\newcommand{\N}{{\mathbb N}}

\newcommand{\cA}{{\mathcal A}}

\newcommand{\cC}{{\mathcal C}}
\newcommand{\cD}{{\mathcal D}}

\newcommand{\cH}{{\mathcal H}}

\newcommand{\cS}{{\mathcal S}}

\newcommand{\cX}{{\mathcal X}}

\usepackage{bm}

\newcommand{\eps}{\varepsilon}

\newcommand{\Exp}{\operatorname*{\mathbf{E}}}
\newcommand{\Ex}{\Exp}

\newtheorem{theorem}{Theorem}[section]

\newtheorem{proposition}[theorem]{Proposition}
\newtheorem{lemma}[theorem]{Lemma}
\newtheorem{corollary}[theorem]{Corollary}

\newtheorem{definition}[theorem]{Definition}

\newtheorem{remark}[theorem]{Remark}

\newcommand{\sdotfill}{\textcolor[rgb]{0.8,0.8,0.8}{\dotfill}} 

\newtheorem{proto}[theorem]{Protocol}
\newtheorem{protoc}[theorem]{Protocol}

\newcommand{\namedref}[2]{#1~\ref{#2}}

\newcommand{\torestate}[3]{%
\expandafter \def \csname BBRESTATE #2 \endcsname{#3}
\theoremstyle{plain}
\newtheorem{BBRESTATETHMNUM#2}[theorem]{#1}
\begin{BBRESTATETHMNUM#2}\label{#2}\csname BBRESTATE #2 \endcsname   \end{BBRESTATETHMNUM#2}
\newtheorem*{BBRESTATETHMNONNUM#2}{\namedref{#1}{#2}}
}

\newcommand{\restate}[1]{\begin{BBRESTATETHMNONNUM#1}[Restated] \csname BBRESTATE #1 \endcsname
\end{BBRESTATETHMNONNUM#1}}

\usepackage{xspace}
\newcommand{\X}{\ensuremath{\mathcal{X}}\xspace} 
\newcommand{\Y}{\ensuremath{\mathcal{Y}}\xspace} 
\renewcommand{\H}{\ensuremath{\mathcal H}\xspace} 

\newcommand{\D}{\cD}
\newcommand{\dist}{D} 

\newcommand{\UUn}{\ensuremath{U_n}\xspace} 

\newcommand{\dis}[1]{\ensuremath{\mathcal{E}\left(#1\right)}\xspace}
\newcommand{\expectedsub}[2]{\ensuremath{\mathop{{}\mathbf{E}}_{#1}\left[#2\right]}\xspace}

\newcommand{\confidence}{\ensuremath{\delta}\xspace}
\newcommand{\Cc}{\ensuremath{\mathcal C}\xspace}
\newcommand{\Ccn}{\ensuremath{{\mathcal C}_n}\xspace}
\newcommand{\concept}{\ensuremath{c}\xspace}

\newcommand{\hypothesisc}{\ensuremath{\mathcal H}\xspace}
\newcommand{\hypoC}{\hypothesisc}

\newcommand{\hconcept}{\ensuremath{h}\xspace}

\newcommand{\entropyfunc}[1]{\ensuremath{H\left(#1\right)}\xspace}

\newcommand{\NN}{\ensuremath{\mathbb N}\xspace}

\newcounter{definitioncnt}
\newcounter{thmcnt}
\newcommand{\assign}{\ensuremath{\leftarrow}\xspace}

\newcommand{\swapping}{\textsc{Swapping Algorithm}\xspace}
\newcommand{\finds}{\textsc{Find-S}\xspace}

\newcounter{prblm}
\newtheorem{prblmdef}[prblm]{Problem Setup}

\begin{document}

\title{\mypapertitle}
\author{\mypaperauthors}
\maketitle

\begin{abstract}
We study adversarial perturbations when the instances are uniformly distributed over $\bits^n$. We study both ``inherent'' bounds that apply to any problem and any classifier 
for such a problem 
as well as bounds that apply to specific problems and specific hypothesis classes.  

As the current literature contains multiple  definitions of adversarial risk and robustness, we start by giving a taxonomy for these definitions based on their direct goals; we identify one of them as the one guaranteeing misclassification by pushing the instances to the \emph{error region}. We then study some classic algorithms for learning monotone conjunctions and compare their adversarial
\iffollowingorders
\else
risk and
\fi
robustness under different definitions by attacking the hypotheses using instances  drawn from the uniform distribution. We observe that sometimes these definitions lead to \emph{significantly different} bounds. Thus, this study advocates for the use of the error-region definition, even though other definitions, in other contexts with context-dependent assumptions, may coincide with the error-region definition.

Using the error-region definition of adversarial perturbations, we then study \emph{inherent} bounds on risk and robustness of \emph{any} classifier for \emph{any} classification problem whose instances are uniformly distributed over $\bits^n$. Using the isoperimetric inequality for the Boolean hypercube, we show that for  initial error $0.01$, there always exists an adversarial perturbation that changes $O(\sqrt{n})$ bits of the instances to increase the risk to $0.5$, making classifier's decisions meaningless. Furthermore, by also using the central limit theorem we show that when $n\to \infty$, at most $c \cdot \sqrt{n}$ bits of perturbations, for a universal constant $c< 1.17$, suffice for increasing the risk to $0.5$, and the same $c \cdot \sqrt{n} $ bits of perturbations \emph{on average} suffice to increase the risk to $1$, hence bounding the robustness by $c \cdot \sqrt{n}$.


 


\end{abstract}


\iffollowingorders
\else
\newpage
\tableofcontents
\fi


\section{Introduction}
In recent years, modern machine learning tools (e.g., neural networks) have pushed to new heights
the classification results on traditional datasets 
that are used as testbeds for various machine learning methods.\setcounter{footnote}{0}\footnote{For example,
\href{http://rodrigob.github.io/are_we_there_yet/build/}{http://rodrigob.github.io/are\_we\_there\_yet/build/}
has a summary of state-of-the-art results.}
As a result, the properties of these methods have been put  into further scrutiny.
In particular, studying the \emph{robustness} of the trained models in various adversarial contexts has gained special attention, leading to the active area of \emph{adversarial} machine learning.

Within adversarial machine learning, 
one particular 
direction of research 
that has gained attention in recent years  
deals with 
the study of the 
so-called \emph{adversarial perturbations} of the test instances.
This line of work was  particularly
popularized, in part, by the work of Szegedy et al.~\citep{Szegedy:intriguing}
 within the context of deep learning classifiers, but the same problem can be asked for general classifiers as well.
%
Briefly, 
when one is given a particular instance $x$ for classification, 
an adversarial perturbation $x'$ for that instance is a new instance with minimal changes in  the features of $x$ so that  the resulting perturbed instance $x'$ is misclassified by the classifier $h$. 
The  perturbed instance $x'$ is commonly referred to as an \emph{adversarial example} (for the classifier $h$). 
Adversarial machine learning has its roots at least as back as in~\citep{Adversarial:Old,Adversarial:models}.
However, the work of~\citep{Szegedy:intriguing} revealed pairs of images that 
differed slightly 
so that a human eye could not
identify any real differences between the two, 
and yet, 
contrary to what one would naturally expect, 
machine learning classifiers would predict different labels for the classifications of such pairs of instances.
It is perhaps this striking resemblance to the human eye of the pairs of images that were provided in~\citep{Szegedy:intriguing} that really gave this new push for intense investigations within the context of adversarial perturbations.
Thus, a very intense line of work started, aiming to understand and explain the properties of machine learning classifiers on such adversarial perturbations; e.g.,~\citep{Adversarial::Harnessing,DeepFool,Adversarial::Constraints,CarliniWagner,madry2017towards}. These attacks are also referred to as \emph{evasion} attacks~\citep{nelson2012query,biggio2014security,Adversarial::Harnessing,CarliniWagner,Adversarial::FeatureSqueezing}. There is also work that aims at making the classifiers more robust under such attacks~\citep{Defenses:Distillation,Adversarial::FeatureSqueezing}, yet newer attacks of Carlini and Wagner~\citep{carlini2017adversarial}   broke many proposed defenses. 


\iffollowingorders
\parag{Our general goal.}
\else
\paragraph{Our general goal.}
\fi
In this work, we study  barriers against  robust classification of adversarial examples.
We are particularly interested in foundational  bounds that potentially apply to broad class of problems and distributions.
One can study this question   from the perspectives  of both risk and robustness. In the case of risk, the adversary's goal is to increase the error probability of the classifier (e.g., to reach risk $0.5$)  by small perturbations of the instances, and in the case of robustness, we are interested in the \emph{average} amount of perturbations needed for making the classifier always fail.



\iffollowingorders
\parag{Studying the uniform distribution.}
\else
\paragraph{Studying the uniform distribution.}
\fi
We particularly study adversarial risk and robustness for learning problems where the input distribution is $U_n$ which is uniform  over the hypercube $\bits^n$. We measure the cost of perturbations using the natural metric of Hamming distance. Namely, the distance between the original and perturbed instances $x,x' \in \bits^n$ is the number of locations that they are different. This class of distributions already include many learning problems of interest.  So, by studying adversarial risk and robustness for such a natural 
distribution, we can immediately obtain results for a broad class of problems. 
We believe it is crucial to understand adversarial risk and robustness  for natural distributions (e.g., $U_n$ uniform over the hypercube) and metrics (e.g., the Hamming distance) to develop a theory of adversarial risk and robustness that can ultimately shed light on the power and limitations of robust classification for practical data sets. Furthermore, natural distributions like $U_n$ model a broad class of learning problems directly; e.g., see
\citep{BlumFJKMR94,SakaiM00,RandomDNF::Uniform,ExactDNF}. The hope is that understanding the limitations of robust learning for these basic natural distributions will ultimately shed light on challenges related to addressing broader problems of interest.

\iffollowingorders
\parag{Related previous work.}
\else
\paragraph{Related previous work.}
\fi
The work of Gilmer et al.~\citep{gilmer2018adversarial} studied the above problem for the special case of input distributions that are uniform over  unit spheres in dimension $n$. They showed that for any classification problem with such input distribution, so long as there is an initial constant error probability $\mu$, the robustness under the $\ell_2$ norm is at most $O(\sqrt{n})$.
 Fawzi et al.~\citep{fawzi2018adversarial} studied the above question for Gaussian distributions in dimension $n$ and showed that when the input distribution has $\ell_2$ norm $\approx 1$, then  by $\approx \sqrt{n}$ perturbations in $\ell_2$ norm, we can make the classifier \emph{change its prediction} (but doing this does not guarantee that the perturbed instance $x'$ will be misclassified). Schmidt et al.~\cite{schmidt2018adversarially} proved limits on  robustness of classifying uniform instances by specific classifiers and using a definition based on ``corrupted inputs'' (see Section  \ref{sec:defs}), while we are mainly interested in  bounds that apply to any classifiers and guarantee misclassification of the adversarial inputs.

\iffollowingorders
\parag{Discussion.}
\else
\paragraph{Discussion.}
\fi
Our negative results of this work, like  other (current proved) bounds in the literature for adversarial risk and robustness only apply
to specific distributions that do not cover the case of distributions that generate images, voices, or other practically interesting data.
We see these results, however, as first steps towards understanding the barriers against robustness. The negative results so far indicate similar phenomena (e.g., relation to isoperimetric inequalities). Thus, as pointed out  in~\citep{gilmer2018adversarial}, these works motivate a deeper study of such inequalities for real data sets. Finally, as discussed in~\cite{fawzi2018adversarial}, such theoretical attacks could \emph{potentially} imply direct attacks on real data, \emph{assuming} the existence of smooth generative models for latent vectors with theoretically nice distributions (e.g., Gaussian or uniform over $\bits^n$) into natural data.

%
%
%

\subsection{Our Contribution and Results}

As mentioned above, our main goal is to understand inherent barriers against robust classification of adversarial examples, and our focus is on the uniform distribution $U_n$ of instances. In order to achieve that goal, we both do a definitions study and prove technical limitation results.

\iffollowingorders
\parag{General definitions and a taxonomy.}
\else
\paragraph{General definitions and a taxonomy.}
\fi
As the current literature contains multiple  definitions of adversarial risk and robustness, we start by giving a taxonomy for these definitions based on their direct goals. More specifically, suppose $x$ is an original instance that the adversary perturbs into a ``close'' instance $x'$. Suppose $h(x),h(x')$ are the predictions of the hypothesis $h(\cdot)$ and $c(x),c(x')$ are the true labels of $x,x'$ defined by the concept function $c(\cdot)$. To call $x'$ a successful ``adversarial example'', a natural definition would compare the predicted label $h(x')$ with some other ``anticipated answer''. However, what $h(x')$ is exactly compared to is where various definitions of adversarial examples diverge. We observe in Section \ref{sec:defs} that the three possible definitions (based on comparing $h(x')$ with either of $h(x)$, $c(x)$ or $c(x')$) lead to three different ways of defining adversarial risk and robustness. 
We then identify one of them (that compares $h(x)$ with $c(x')$) as the one guaranteeing misclassification by pushing the instances to the \emph{error region}. We also discuss natural conditions under which these definitions coincide.
However, these conditions do not hold \emph{in general}.

\iffollowingorders
\parag{A comparative study through monotone conjunctions.}
\else
\paragraph{A comparative study through monotone conjunctions.}
\fi
We next ask: how close/far are these definitions in settings where, e.g., the instances are drawn from the uniform distribution? To answer this question, we make a  comparative study of adversarial risk and robustness for a particular case of learning monotone conjunctions  under the uniform distribution $U_n$ (over $\bits^n$). 
A monotone conjunction $f$ is a function of the form 
$f=(x_{i_1} \wedge \dots \wedge x_{i_k})$.
This class of functions is perhaps one of the most natural and basic learning problems 
that are studied in computational learning theory
as it encapsulates, in the most basic form, the class of functions that determine
which features should be included as relevant for a prediction mechanism.
For example, 
Valiant in~\citep{Valiant::Evolvability} used this class of functions under \UUn to exemplify the framework of evolvability. 
We 
attack monotone conjunctions under \UUn 
in order to 
contrast  different behavior of definitions of adversarial risk and robustness.

In Section \ref{sec:mon-conj}, we show that previous definitions of robustness that are not based on the error region, lead to bounds that do \emph{not} equate the  bounds provided by the error-region approach.
We do so by first deriving theorems that characterize the adversarial risk and robustness of a given hypothesis and a concept function under the uniform distribution. Subsequently, by performing experiments  we show that, on average, hypotheses computed by two popular algorithms (\finds~\citep{Book::Mitchell:MachineLearning} and \swapping~\citep{Valiant::Evolvability}) also exhibit the behavior that is predicted by the theorems. Estimating the (expected value of) the adversarial risk and robustness of hypotheses produced by \emph{other} classic algorithms under specific distributions, or for other concept classes, is an interesting future work.

\iffollowingorders
\parag{Inherent bounds for any classification task under the uniform distribution.}
\else
\paragraph{Inherent bounds for any classification task under the uniform distribution.}
\fi
Finally, after establishing further motivation to use the error-region definition as the default definition for studying adversarial examples in \emph{general} settings, we  turn into studying \emph{inherent} obstacles against robust classification when the instances are drawn from the uniform distribution.
We prove that for \emph{any} learning problem $\problem$ with input distribution $U_n$ (i.e., uniform over the hypercube) and for any classifier $h$ for $\problem$ with a constant error $\mu$, the  robustness of $h$ to adversarial perturbations (in Hamming distance) is  at most $O(\sqrt n)$.  We also show that by the same amount of $O(\sqrt n)$ perturbations \emph{in the worst case}, one can increase the risk to $0.99$. Table~\ref{tbl:violate} lists some numerical examples.
\remove{, rendering binary classifiers' decisions essentially meaningless.\footnote{That is because in a binary classification, a random decision is correct with probability $0.5$.}  Our experimental calculations for concrete values of $n$ in the range of $n=10^3$ to $n=10^5$ indicate that even a better bound of $1.17 \sqrt n$ holds for both scenarios of adversarial risk and robustness, and the bound gets even smaller for larger values of $n$. In concrete terms, for any learning problem with test distributions $U_{10^4}$, and any classifiers for that problem with initial error $0.01$, perturbing only $117$ locations (in the worst case) is enough to increase the risk from $0.01$ to $0.5$ and $117$ perturbations \emph{on average} are sufficient to increase the risk basically to $1$ (i.e., making predictions to be always wrong). 
Table~\ref{tbl:violate} lists some numerical examples.
}
\begin{table}[ht!]
    \centering
    \caption{Each row focuses on the number of tampered bits  to achieve its stated goal. The second column shows  results using direct calculations for specific dimensions. The third column shows that these results are indeed achieved in the limit, and the last column shows bounds proved for all $n$.
    }
    \label{tbl:violate}
    \begin{tabular}{|c||c|c|c|}\cline{2-4}
   \multicolumn{1}{c|}{} & \multicolumn{3}{c|}{Types of bounds} \\\hline
         Adversarial goals & $n=10^3,10^4,10^5$       & $n \To \infty$       & all $n$         
         \\\hline\hline
            From initial risk $0.01$ to $0.99$ & $\approx 2.34 \sqrt n$         & $<2.34 \sqrt n$        & $<3.04 \sqrt n$ 
            \\\hline
                 From initial risk $0.01$ to $0.50$ & $\approx 1.17 \sqrt n$         & $< 1.17 \sqrt n$        & $<1.52 \sqrt n$ 
            \\\hline
    Robustness  for  initial risk $0.01$       & $\approx 1.17 \sqrt n$   & $< 1.17 \sqrt n$  & $<1.53 \sqrt n$   
    \\\hline
    \end{tabular}
\end{table}

\remove{Our results are \emph{information theoretic}, namely, we only show the \emph{existence} of adversarial examples. This already shows that no  defense exists that provably beats our bounds.
}
To prove results above, we apply the isoperimetric inequality of~\citep{isoperiHQ,harper1966optimal} to the error region of the classifier $h$ and the ground truth $c$. In particular, it was shown in~\citep{harper1966optimal,isoperiHQ} that the subsets of the hypercube with minimum ``expansion'' (under Hamming distance) are Hamming balls. This fact enables us to prove our bounds on the risk. We then prove the bounds on robustness by proving a general connection between risk and robustness that might obe of independent interest.  Using the central limit theorem, we sharpen our bounds for robustness and obtain bounds that closely match the bounds that we also obtain by direct calculations (based on the isoperimetric inequalities and picking Hamming balls as error region) for specific values of dimension $n=10^3,10^4,10^5$.
\remove{
So, if consider the ``expanded'' error region that comes as the result of adversary's tampering with input instances, the minimum  number of adversarial instances occur when the original error region is of the shape of a ball; namely, any other geometry for the initial error region, the adversarial risk would be \emph{more} under adversarial tampering of the instances. Now, we can focus on  error regions that of the shape of a Hamming ball, in which case we can  calculate the asymptotic as well as exact numerical bounds for adversarial risk and robustness (Table~\ref{tbl:violate}) that hold for \emph{any} classifiers. See Section~\ref{sec:uniform} for details of the proofs.
}

\iffollowingorders
\parag{Full version.}
{All proofs could be found in the full version of the paper, which also includes results related to the adversarial risk of monotone conjunctions, complementing the picture of Section~\ref{sec:mon-conj}.}
\else
\fi

\section{General Definitions of Adversarial Risk and Robustness} \label{sec:defs}

\paragraph{Notation.} 
We  use calligraphic letters (e.g., $\cX$) for sets and capital non-calligraphic letters (e.g., $D$) for distributions. 
By $x \gets D$ we denote sampling $x$  from $D$. 
In a classification problem $\problem=(\X,\Y,\D,\C,\H)$, the set $\X$ is the set  of possible \emph{instances}, 
\Y is the set of possible \emph{labels}, 
$\D$ is a set of distributions over $\X$,  
$\C$ is a class of \emph{concept} functions, and $\H$ is a class of \emph{hypotheses}, where any $f\in\C \cup \cH$ is a mapping from $\X$ to $\Y$.
An \emph{example} is a \emph{labeled instance}.
We did not state the loss function explicitly, as we work with classification problems, however all main three definitions of this section directly extend to arbitrary loss functions.
For  $x \in \X, c \in \C, \dist \in \D$,
the \emph{risk} or \emph{error} of a hypothesis $h \in \H$ is the expected ($0$-$1$) loss of $(h,c)$ with respect to $\dist$, namely $\Risk(h,c,\dist) = \Pr_{x \gets\dist}[h(x) \neq c(x)]$. 
We are usually interested in learning problems  with a fixed distribution $\D=\set{\dist}$, as we are particularly interested in  robustness of learning under the uniform distribution $U_n$ over $\bits^n$. Note that since we deal with negative results, fixing the distribution only makes our results stronger. As a result, whenever $\D=\set{\dist}$, we omit $\dist$ from the risk notation and simply write  $\Risk(h,c)$.
We usually work with problems $\problem=(\X,\Y,\dist,\C,\H,\metric)$ that include a metric $\metric$ over the instances. For a set $\cS \se \X$ we let $\metric(x,\cS) = \inf \set{\metric(x,y) \mid y \in \cS}$ and  $\Ball_r(x)=\set{x' \mid \metric(x,x') \leq r}$. By $\HD$ we denote Hamming distance for pairs of instances from $\bits^n$.  
Finally, we use the term \emph{adversarial instance} to refer to an adversarially perturbed instance  $x'$ of an originally sampled instance $x$ when the label of the adversarial example is either not known or not considered.

\iffollowingorders
\else
\subsection{Different Definitions and Qualitative 
Comparisons}\label{sec:definitions:perturbations}
\fi

Below we present  formal definitions of adversarial risk and robustness.
In all of these  definitions we will deal with attackers who  perturb the initial test instance $x$ into a \emph{close} adversarial instance $x'$. We will measure how much an adversary can increase the \emph{risk} by  perturbing a given input $x$ into a \emph{close} adversarial example $x'$. These definitions differ in when to call $x'$ a successful adversarial example. First we formalize the main definition that we use in this work based on adversary's ability to push instances to the error region.

\begin{definition}[Error-region  risk and robustness] \label{def:ER}
Let $\problem=(\X,\Y,\dist,\C,\H,\metric)$ be a classification problem  (with metric $\metric$ defined over instances $\X$).
\begin{itemize}
    \item {\bf Risk.}  For  any $r \in \Rplus, h \in \H, {c \in \C}$, the \emph{error-region  risk} under $r$-perturbation is
    \begin{displaymath}
    \Risk_r\ER(h,{c}) = \Pr_{x \gets \dist}[\exists x' \in \Ball_r(x), h(x') \neq \textcolor{red}{{c(x')}}]\,.
    \end{displaymath}
    For $r=0$, $\Risk_r\ER(h,{c})=\Risk(h,{c})$ becomes the standard notion of risk.
    \item {\bf Robustness.} For any $h \in \H, x\in \X, {c\in \C}$, the 
     \emph{error-region  robustness} is the expected distance of a sampled instance to the error region, formally defined as follows
         \begin{displaymath}
    \Rob\ER(h,c) = \expectedsub{x \gets \dist}{\inf \set {r \colon \exists x' \in \Ball_r(x), h(x') \neq \textcolor{red}{c(x')}}}\,.
    \end{displaymath}
\end{itemize}
\end{definition}
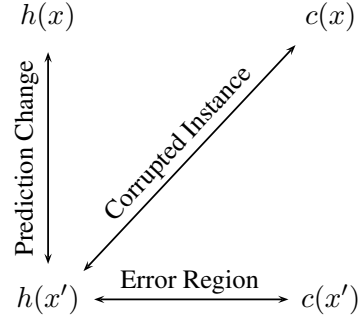
\begin{wrapfigure}{r}{0.5\textwidth}
{
\iffollowingorders
\vspace*{-0.6cm}
\else
\fi
\centering
\begin{pspicture*}(0,0)(10,10)
\rput(1,9){$h(x)$}
\rput(1,1){$h(x')$}
\rput(9,9){$c(x)$}
\rput(9,1){$c(x')$}
\psline[linecolor=black]{<->}(2.3,1)(7.8,1)   
\psline[linecolor=black]{<->}(1,2)(1,8)        
\psline[linecolor=black]{<->}(2,1.8)(8,8.2)  
%
%
%
\rput(5,1.5){\small Error Region}             %
\rput{47}(4.7,5.4){\small Corrupted Instance} %
\rput{90}(0.4,5){\small Prediction Change}    %
\end{pspicture*}
\iffollowingorders
\vspace*{-0.19cm}
\else
\fi
\caption{The three main definitions based on what $h(x')$ is compared with. 
}\label{fig:explanation}
}
\end{wrapfigure}
Definition \ref{def:ER} requires the adversarial instance $x'$ to be \emph{misclassified}, namely, $h(x') \neq c(x')$. So, $x'$ clearly belongs to the error region of the hypothesis $h$ compared to the ground truth $c$. 
This definition is implicit in the work of~\cite{gilmer2018adversarial}. 
%
In what follows, we compare  our main definition above with previously proposed definitions of adversarial risk and robustness found in the literature and discuss when they are (or when they are not) equivalent to Definition \ref{def:ER}. 
Figure \ref{fig:explanation} summarizes the differences between the three main definitions that have appeared in the literature, where we distinguish cases by comparing the classifier's prediction $h(x')$ at the new point $x'$ with either of $h(x)$, $c(x)$, or $c(x')$, leading to three different definitions.


\paragraph{Definitions based on hypothesis's prediction change (PC risk and robustness).} 
Many works, including the works of \citep{Szegedy:intriguing,fawzi2018adversarial}  use a definition of robustness that compares classifier's prediction $h(x')$ with the prediction $h(x)$ on the original instance $x$. Namely, they require $h(x') \neq h(x)$ rather than $h(x') \neq c(x')$ 
in order to consider $x'$ an adversarial instance.
Here we refer to this definition (that does not depend on the ground truth $c$) as \emph{prediction-change} (PC) risk and robustness (denoted as  $\Risk_r\PC(h) $ and $\Rob\PC(h) $). We note that this definition captures the error-region risk and robustness if we \emph{assume} the initial correctness (i.e., $h(x)=c(x)$) of classifier's prediction on all $x \gets X$ and ``truth proximity'', i.e., that   $c(x) = c(x')$ holds for all $x'$ that are ``close'' to $x$. Both of these  assumptions are valid in some natural scenarios. For example, when input instances  consist of images that look similar to humans (if used as the ground truth $c(\cdot)$) and if $h$ is also correct on the original (non-adversarial) test examples, then  the two definitions (based on error region or prediction change) coincide. But, these assumptions do not hold in \emph{in general}.


\iffollowingorders
\else
We note that there is also a work in the direction of finding adversarial instances that may potentially fool humans that have limited time to decide for their label, as in \citep{Adversarial:TimeLimited}. The images of \citep{Adversarial:TimeLimited} are sufficiently `confusing' that answers of the form ``I do not know'' are very plausible from the humans that are asked. This fuzzy classification that allows ``I do not know'' answers is reminiscent of the \emph{limited membership query} model of Sloan and Tur{\'a}n \citep{LimitedMembership} (which is a worst-case version of the \emph{incomplete membership query} model of Angluin and Slonim \citep{IncompleteMembership}; 
see also~\citep{AMST:MaliciousOmissionsErrors} and \citep{QueriesBookChapter} for further related discussions) as well as of the model of learning from a \emph{consistently ignorant teacher} of Frazier et al.~\citep{IgnorantTeacher}.
\fi

\paragraph{Definitions based on the notion of corrupted instance (CI risk and robustness).} The works of~\citep{mansour2015robust,feige2015learning,feige2018robust,attias2018improved} study the robustness of learning models in the presence of \emph{corrupted inputs}.  A more recent framework was developed in~\citep{madry2017towards,schmidt2018adversarially} for modeling risk and robustness that is inspired by robust optimization~\citep{BEN:09} (with an underlying metric space) and model adversaries that corrupt the the original instance in (exponentially more) ways. 
When studying adversarial perturbations using corrupted instances, we define adversarial risk by requiring  the adversarial instance $x'$ to satisfy $h(x') \neq c(x)$. The term ``corrupted instance''  is particularly helpful as it emphasizes on the fact that the goal (of the classifier) is to find the \emph{true} label of the \emph{original} (uncorrupted) instance $x$, while we are only given a corrupted version $x'$. 
Hence, we refer to this definition as the \emph{corrupted instance} (CI) risk and robustness and denote them by $\Risk_r\CI(h,{c}) $ and $\Rob\CI(h,{c}) $. 
The advantage of this definition compared to the prediction-change based definitions is that here, we no longer need to assume the initial correctness assumption.  Namely, only if the ``truth proximity'' assumption holds, then we have $c(x)=c(x')$ which together with the condition $h(x') \neq c(x)$ we can conclude that $x'$ is indeed misclassified. However, if small perturbations can change the ground truth, $c(x')$ can be different from $c(x)$, in which case, it is no long clear whether $x'$ is misclassified or not. 

\paragraph{Stronger definitions  with more restrictions on adversarial instance.} The corrupted-input definition requires an adversarial instance $x'$ to satisfy $h(x') \neq c(x)$, and the error-region definition requires $h(x') \neq c(x')$. What if we require \emph{both} of these conditions to call $x'$ a true adversarial instance? This is indeed the definition used in the work of Suggala et al.~\citep{strong}, though more formally in their work, they subtract the original risk (without adversarial perturbation) from the adversarial risk. This definition is certainly a \emph{stronger} guarantee for the adversarial instance. Therefore, we simply refer to risk and robustness under this condition as \emph{strong} adversarial risk and robustness. As this definition is a hybrid of the error-region and corrupted-instance definitions, we do not make a direct study of this definition and only focus on the other three definitions described above.

\paragraph{How about when the classifier $h$ is 100\% correct?} We emphasize that when $h$ happens to be the same function as $c$, (the error region) Definition \ref{def:ER} implies $h$ has zero \emph{adversarial}
risk and infinite adversarial robustness $\Rob\ER(h,c)=\infty$. This is expected, as there is no way an adversary can perturb any input $x$ into a misclassified $x'$. However, both of the definitions of risk and robustness based on  prediction change~\citep{Szegedy:intriguing} and corrupted instance~\citep{mansour2015robust,madry2017towards} could compute large risk and small robustness for such $h$. In fact, in a recent work
\citep{Tsiprasetal:corrupted:at_odds_with_accuracy} it is shown that for definitions based on corrupted input, correctness might be \emph{provably at odds} with robustness in some cases.
Therefore, even though all these definitions could perhaps be used to approximate the risk and robustness when we do not have access to the ground truth $c'$ on the new point $x'$, in this work  we separate the \emph{definition} of risk and robustness  from how to compute/approximate them, so we will use Definition \ref{def:ER} by default.

\iffollowingorders
\else
\subsection{Various Aspects of the Attack Models}\label{sec:efficient-attacks}
We emphasize that the definitions of Section \ref{sec:definitions:perturbations} are all \emph{information theoretic} and do not address the \emph{efficiency}  of the adversary who perturbs the original instance $x$ into $x'$. 
Moreover, there are other aspects of the attack that are implicit in the definitions Section \ref{sec:definitions:perturbations} in terms of what adversary does or does not have access to during the course of the attack.
Below, we briefly point out these other aspects.

\begin{itemize}
    \item {\bf Efficiency.} This aspect of an attack could come in two flavor. One way to mathematically formalize ``efficient'' attacks is to use polynomial-time attacks as it is done in cryptography. Another way is to use information theoretic attacks without the efficiency requirements. Security against information theoretic attacks are stronger, while attacks of polynomial-time form are stronger.
    
    \item {\bf Information access.} The other aspect of the attack is about \emph{what adversary has access to} during the attack and \emph{how it can access} this information. We separate thes aspects as follows.
    \begin{itemize}
        \item {\bf What to access.} In general, we can consider attacks that do or do not access to either of the ground truth $c$, the hypothesis $h$, or  distribution $D$.
        \item {\bf How to access.} If the attack can depend on a function $f$ (e.g., $f=h$ or $f=c$) or a distribution $D$ it can still access this information in various forms. An information theoretic attack can completely depend on the full description of $f$, while an efficient (polynomial time attack) can use oracle access to $f$ (regardless of efficiency of $f$ itself) or a sampler for $D$. In fact, if $f$ (or a sampler for a distribution $D)$ has a compact representation, then an efficient attacker can also fully depend on $f$ or $D$ if that representation is given.
    \end{itemize}
\end{itemize}

Going back to the definitions of Section \ref{sec:defs}, by  ``$\exists x' \in\Ball_r(x), P(x')$'' we simply state the \emph{existence} of a close instance $x'$ with a property $P(x')$ while it might be computationally infeasible to actually \emph{find} such an $x'$. Moreover, the definitions of Section \ref{sec:defs} assume the adversary has full access to $f,c,D$.

\fi

\section{A Comparative Study through Monotone Conjunctions}\label{sec:mon-conj}
In this section, we compare the risk and robustness under the three definitions of Section \ref{sec:defs} through a study of  monotone conjunctions under the uniform distribution.  Namely, we consider adversarial perturbations of truth assignments 
that are drawn from the uniform distribution \UUn over $\{0, 1\}^n$
when the concept class contains  monotone conjunctions.  
As we will see, these definitions diverge in this natural case.
Below we fix the setup under which all the subsequent results are obtained.
\begin{prblmdef}\label{setup:mon-conj}
Let \Ccn be the concept class of all 
monotone conjunctions formed by 
at least one and at most $n$ Boolean variables. 
The target concept (ground truth) $c$ that needs to be learned is drawn from \Ccn. Let the hypothesis class 
be $\hypothesisc = \Ccn$ and let  $\hconcept\in\hypothesisc$ be the hypothesis obtained by a learning algorithm after processing the training data.
With $\abs{h}$ and $\abs{c}$ we denote the \emph{size} of $h$ and $c$ respectively; that is, number of variables that $h$ and $c$ contain.\footnote{
For example, $h_1 = x_1\wedge x_5\wedge x_8$ is a monotone conjunction of three variables in a space where we have $n\ge 8$ variables and $\abs{h_1} = 3$.}
Now let, 
\begin{equation}\label{eq:form}
\concept = \bigwedge_{i=1}^m x_i \wedge\bigwedge_{k=1}^u y_k
\,\,\,\,\,\,\,\,\,\,\,\,\,
\,\,\,\,\,\,\,\,\,\,\,\,\,
\mbox{and} 
\,\,\,\,\,\,\,\,\,\,\,\,\,
\,\,\,\,\,\,\,\,\,\,\,\,\,
\hconcept = \bigwedge_{i=1}^m x_i \wedge\bigwedge_{\ell=1}^w z_\ell\,. 
\end{equation}
We will call the variables that appear both in \hconcept and \concept as \emph{mutual},
the variables that appear in \concept but not in \hconcept as \emph{undiscovered},
and the variables that appear in \hconcept but not in \concept as \emph{wrong} (or \emph{redundant}).
Therefore in (\ref{eq:form}) we have $m$ mutual variables, $u$ undiscovered and $w$ wrong. We denote the error region of a 
hypothesis \hconcept and the target concept \concept
with $\dis{\hconcept, \concept}$. 

That is,
$\dis{\hconcept, \concept} = \set{x\in\{0, 1\}^n \mid \hconcept(x) \neq \concept(x)}$.
%
%
The probability mass of the error region between $h$ and $c$, denoted by $\mu$, under the uniform distribution \UUn over $\bits^n$ is then,
%
\begin{equation}\label{eq:dis-region:count}
\Pr_{x \gets U_n}[x \in \dis{h,c}] = \mu = (2^{w} + 2^u - 2)\cdot 2^{-m-u-w} \,.
\end{equation}
In this problem setup we are interested in computing the \emph{adversarial risk} and \emph{robustness} that attackers can achieve when instances are drawn from the uniform distribution \UUn over $\bits^n$.
\end{prblmdef}

\begin{remark}
Note that $\mu$ is a variable that depends on the particular $h$ and $c$.
\end{remark}

\iffollowingorders
\else
\begin{remark}\label{rem:mon-conj:efficient-attack}
Connecting to our discussion from Section \ref{sec:efficient-attacks}, adversaries who have 
oracle access to such hypotheses 
can identify the variables that appear in them and so reconstruct them efficiently as follows. Since 
$x_{one} = \langle 1, 1, 1, \ldots, 1\rangle$ is always positive one can query all the $n$ vectors that have all $1$'s but one $0$, and thus determine all the variables that appear in a particular hypothesis.
\end{remark}
\fi

\iffollowingorders
Using the Problem Setup \ref{setup:mon-conj}, 
in what follows we compute the adversarial robustness that an arbitrary hypothesis has against an arbitrary target using the \emph{error region (ER)} definition that we advocate in contexts where the perturbed input is supposed to be misclassified and do the same calculations for adversarial risk and robustness that are based on the definitions of \emph{prediction change (PC)} and \emph{corrupted instance (CI)}. The important message is that the adversarial robustness of a hypothesis based on the ER definition is $\Theta\left(\min\{\abs{h}, \abs{c}\}\right)$, whereas the adversarial robustness based on PC and CI is $\Theta\left(\abs{h}\right)$.
In the full version of the paper we also give theorems (that have similar flavor) for calculating the adversarial risk based on the three main definitions (ER, PC, CI).
\else
Using the Problem Setup \ref{setup:mon-conj}, 
in what follows we compute the adversarial risk and robustness that an arbitrary hypothesis has against an arbitrary target using the \emph{error region (ER)} definition that we advocate in contexts where the perturbed input is supposed to be misclassified and do the same calculations for adversarial risk and robustness that are based on the definitions of \emph{prediction change (PC)} and \emph{corrupted instance (CI)}.
The adversarial robustness of a solution using the prediction change and corrupted instance definitions 
is proportional to the size (number of variables) of the learned solution; 
experimentally we obtain that it is proportional to about \emph{half} the number of variables that the hypothesis contains.
On the other hand, 
the adversarial robustness of a learned solution using the error region definition
is proportional to the \emph{minimum} between the size of the hypothesis and the size of the target;
experimentally we obtain that it is proportional to about \emph{half} that minimum value.
In other words, for a hypothesis that has many variables and a target that has fairly few variables, 
the prediction change and corrupted instance definitions imply large robustness for the learned solution, whereas the adversarial robustness as implied by the error region definition is low.
This last setup is precisely the case for a large set of target functions when PAC learning
monotone conjunctions with 
the \finds algorithm;
see Section \ref{sec:experiments:conjunctions}
\fi

\iffollowingorders
\else
\subsection{Error Region Risk and Robustness}
\begin{restatable}[Error region risk - lower bound]{theorem}{ThmDisRisk}\label{thm:dis:risk}
Consider the Problem Setup \ref{setup:mon-conj}.
Then, if $\hconcept = \concept$, we have 
$\Risk\nomDR(h,c) = 0$, while if $\hconcept \neq \concept$, 
then, with a perturbation budget of $r$ we can obtain the following lower bounds. 

\begin{itemize}
\item If $0 \le r \le \lfloor m/2\rfloor$, then 
$\Risk\nomDR(h,c) \ge \mu\cdot\sum_{j=0}^r \binom{m}{j}$, where $\mu$ is given by (\ref{eq:dis-region:count}).
\item If $\lfloor\frac{m}{2}\rfloor + 1 \le r = \lfloor\frac{m}{2}\rfloor + \gamma \le \lfloor\frac{m}{2}\rfloor + \lfloor\frac{\min\{u, w\}}{2}\rfloor$, then\\
$$\Risk\nomDR(h,c) \ge \frac{1}{4}\cdot 2^{-\min\{u, w\}} \cdot \sum_{\zeta=1}^{\gamma}\binom{\min\{u, w\}}{\zeta}.$$
\item If 
$\lfloor\frac{m}{2}\rfloor + \lfloor\frac{\min\{u, w\}}{2}\rfloor + 1 \le r \le \min\set{\abs{h}, \abs{c}}$,
then
$\Risk\nomDR(h,c) \ge \frac{1}{8}$. 
\item If  $1 + \min\set{\abs{h}, \abs{c}} \le r$, then 
$\Risk\nomDR(h,c) = 1$.
\end{itemize}
\end{restatable}
\iffollowingorders
\else
\begin{proof}[Proof Sketch]
Assume that $h\neq c$, since otherwise the risk is $0$.

We distinguish cases for the various values that $m, u, w$ can take in  (\ref{eq:form}):
\begin{inparaenum}[(i)]
\item $m=0$, 
\item $m\ge 1$ and $u = 0$ and $w \ge 1$, 
\item $m \ge 1$ and $u \ge 1$ and $w = 0$, 
and finally, 
\item \textbf{the more general case, where \bm{$m\ge 1$}, \bm{$u \ge 1$} and \bm{$w\ge 1$}.}
\end{inparaenum}
Below we will prove fully the more involved case, \textbf{case (iv)};
the other cases can easily be obtained using (iv) as a guide. Furthermore, we distinguish between two main cases: having a budget $r \le \min\set{\abs{h}, \abs{c}}$ versus having a budget of $r > \min\set{\abs{h}, \abs{c}}$.


\iffollowingorders
{\bf Case 1: Budget $0 \le r \le \min\{\abs{\hconcept}, \abs{\concept}\}$.}
\else
\paragraph{Case 1: Budget \bm{$0 \le r \le \min\{\abs{h}, \abs{c}\}$}.}
\fi
We distinguish cases based on the relationship between the prediction and the true label of a randomly drawn instance.
\iffollowingorders
\begin{asparadesc}
\else
\begin{description}
\fi
\item [Case 1A: Budget \bm{$0 \le r \le \min\{\abs{h}, \abs{c}\}$} and instance \bm{$x$} such that \bm{$h(x) \neq c(x)$}.]
When $h$ and $c$ disagree, this means that 
$x\in\dis{h, c}$ 
and hence without the need of any budget (i.e., $r = 0$) 
all these truth assignments contribute to the risk of the hypothesis.
By (\ref{eq:dis-region:count})
we obtain such an $x$ 
with probability 
$(2^{w} + 2^u - 2)\cdot 2^{-m-u-w} = \mu =  \mu\cdot\binom{m}{0}$.

\item [Case 1B: Budget \bm{$0 \le r \le \min\{\abs{h}, \abs{c}\}$} and instance \bm{$x$} such that \bm{$h(x) = c(x) = 1$}.]
With probability $2^{-m-u-w}$, $x$ satisfies both $c$ and $h$. 
Since $h\neq c$ there is at least one variable that appears in either $c$ or $h$ but not in both. 
Therefore, with a budget of $r\ge 1$ we can flip that one bit in $x$ that corresponds to that variable and the resulting $x'$ will be misclassified. 

\item [Case 1C: Budget \bm{$0 \le r \le \min\{\abs{h}, \abs{c}\}$} and instance \bm{$x$} such that \bm{$h(x) = c(x) = 0$}.]
With probability 
$1 - \mu - 2^{-m-u-w}$, 
$x$ falsifies both $c$ and $h$.
%
We 
distinguish cases further based on the range of the budget $r$ that is provided to us.


\begin{description}
\item [\textbf{Case 1C1: \bm{$0 \le r \le \lfloor m/2\rfloor$}.}]
We look at truth assignments that have 
$1\le j \le r \le \lfloor m/2\rfloor$ 0's among the $m$ mutual variables
and further have: 
\emph{(i)} $1\le \zeta \le u$ undiscovered variables falsified 
and all the $w$ wrong variables satisfied, and
\emph{(ii)} all the $u$ undiscovered variables satisfied 
and $1\le \xi \le w$ wrong variables falsified.
With a budget of $r$, 
the contribution to the adversarial risk by these 
assignments 
is, 
\begin{displaymath}
2^{-m-u-w}\sum_{j=1}^r \binom{m}{j}\left[\binom{w}{0}\sum_{\zeta=1}^{u} \binom{u}{\zeta} 
+ \binom{u}{0}\sum_{\xi=1}^{w}\binom{w}{\xi}\right]
\end{displaymath}
which is, 
$
\left[2^{-m-u} + 2^{-m-w} - 2^{1-m-u-w}\right]\sum_{j=1}^r \binom{m}{j}$
and by (\ref{eq:dis-region:count})
it is 
$\mu\cdot\sum_{j=1}^r \binom{m}{j}$.

\item 
[\textbf{Case 1C2: \bm{$\lfloor\frac{m}{2}\rfloor + 1 \le r  
\le \lfloor\frac{m}{2}\rfloor + \lfloor\frac{\min\{u, w\}}{2}\rfloor$}.}]
Assume\footnote{The opposite case where $w \le u$ is symmetric. Also, $u, w > 0$ since we explore case (iv), the more general case for the proof, where $m \ge 1$, $u \ge 1$ and $w \ge 1$.} that $1 \le u \le w \Rightarrow \abs{\concept}\le \abs{\hconcept}$. 
We look at the truth assignments where we have at most $\lfloor m/2\rfloor$ 0's among the $m$ mutual variables and further we have $\gamma$ 0's where $\gamma \ge 1$ among the $u$ undiscovered variables plus at least one 0 among the $w$ wrong variables.
With a budget of $r = \lfloor\frac{m}{2}\rfloor + \gamma$ 
we flip the 0's that exist among the mutual variables 
plus the $\gamma$ undiscovered variables that are currently falsified in $x$
and thus we hit the error region.
Therefore, the contribution to the adversarial risk by these truth assignments alone is, 
\begin{displaymath}
2^{-m-u-w}
\sum_{j=0}^{\lfloor m/2\rfloor} \binom{m}{j}\sum_{\xi=1}^{w}\binom{w}{\xi}\sum_{\zeta=1}^{\gamma}\binom{u}{\zeta}
\end{displaymath}
which is, 
$
%
\frac{1}{2} \cdot 
(1-2^{-w}) \cdot 
2^{-u} \cdot 
\sum_{\zeta=1}^{\gamma}\binom{u}{\zeta} 
\ge 
\frac{1}{4} \cdot
2^{-u} \cdot 
\sum_{\zeta=1}^{\gamma}\binom{u}{\zeta}$.
(In the opposite case, where $1 \le w < u$ we obtain a lower bound of 
$\frac{1}{4} \cdot
2^{-w} \cdot 
\sum_{\zeta=1}^{\gamma}\binom{u}{\zeta}$
that explains the statement of the theorem.)

\item 
[\textbf{Case 1C3: \bm{$\lfloor\frac{m}{2}\rfloor + \lfloor\frac{\min\{u, w\}}{2}\rfloor + 1 \le r \le m + \min\{u, w\}$}.}]
Assume that $u\le w$. (Again, the opposite case is symmetric.)


We look at the truth assignments where we have 
at most $\lfloor m/2\rfloor$ many $0$'s among the $m$ mutual variables
and further we have at most $1+\lfloor u/2\rfloor$ many  $0$'s among the 
$u$ undiscovered variables 
plus at least one $0$ among the $w$ wrong variables.
With a budget of $r \ge \lfloor\frac{m}{2}\rfloor + \lfloor u/2\rfloor + 1$ 
we flip the $0$'s that exist among the mutual variables 
plus the $0$'s that exist among the undiscovered variables
and thus we hit the error region.
Therefore, the contribution to the adversarial risk by these truth assignments alone is, 
\begin{displaymath}
2^{-m-u-w}
\sum_{j=0}^{\lfloor m/2\rfloor} \binom{m}{j}
\sum_{\zeta=1}^{1+\lfloor u/2\rfloor}\binom{u}{\zeta}
\sum_{\xi=1}^{w}\binom{w}{\xi}\,.
\end{displaymath}
By Lemma~\ref{lem:ratio-at-least-half}, 
$\sum_{j=0}^{\lfloor m/2\rfloor} \binom{m}{j} \ge 2^{m-1}$
as well as 
$\sum_{\zeta=1}^{1+\lfloor u/2\rfloor}\binom{u}{\zeta} \ge \sum_{\zeta=0}^{\lfloor u/2\rfloor}\binom{u}{\zeta} \ge 2^{u-1}$,
we obtain the lower bound, 
$
2^{-m-u-w}\cdot 2^{m-1}\cdot 2^{u-1}\cdot (2^{w}-1) = 
\frac{1}{4}\cdot(1 - 2^{-w}) \ge \frac{1}{8} 
$.
%
%
%
%
\end{description}
\iffollowingorders
\end{asparadesc}
\else
\end{description}
\fi


\iffollowingorders
{\bf Case 2: Budget $r\ge 1 + \min\{\abs{\hconcept}, \abs{\concept}\}$.}
\else
\paragraph{Case 2: Budget \bm{$r\ge 1 + \min\{\abs{h}, \abs{c}\}$}.}
\fi
The risk 1 since we can hit the error region by making at most 
$1 + \min\{\abs{\hconcept}, \abs{\concept}\}$ changes in any given truth assignment.
This worst case scenario can be observed when $h$ is a specialization of $c$ (or vice versa)
and in particular for truth assignments where all the m mutual variables are falsified
as well as all the wrong (resp., undiscovered) variables are satisfied.
Then, we need to flip the $m = \min\{\abs{h}, \abs{c}\}$ mutual variables plus one more among the
wrong (resp., undiscovered) in order to hit the error region.
\end{proof}
\fi
\fi

\iffollowingorders
\begin{restatable}{theorem}{ThmDisRobustness}\label{thm:dis:robustness}
Consider the Problem Setup \ref{setup:mon-conj}.
Then, if $\hconcept = \concept$ we have $\Rob\nomtDR(h, c) = \infty$, while if $\hconcept \neq \concept$ we have $\min\{\abs{h}, \abs{c}\}/16 \le \Rob\nomtDR(h, c) \le 1 + \min\{\abs{h}, \abs{c}\}$.
\end{restatable}
\else
\begin{restatable}[Error region robustness]{theorem}{ThmDisRobustness}\label{thm:dis:robustness}
Consider the Problem Setup \ref{setup:mon-conj}.
Then, $\Rob\nomtDR(h, c) = \infty$ when $\hconcept = \concept$, while if $\hconcept \neq \concept$ we have,  
\begin{displaymath}
\frac{1}{16}\cdot\min\{\abs{h}, \abs{c}\} \le \Rob\nomtDR(h, c) \le 1 + \min\{\abs{h}, \abs{c}\}.
\end{displaymath}
\end{restatable}
\fi
\iffollowingorders
\else
\begin{proof}[Proof Sketch]
$\hconcept = \concept \Rightarrow \Rob\nomtDR(h, c) = \infty$. 
Hence, below we will examine the case where $\hconcept \neq \concept$.

As in Theorem \ref{thm:dis:risk} 
below we prove fully case (iv) where $m\ge 1$, $u \ge 1$ and $w\ge 1$. Using the analysis that we present below for case (iv) as a guide, we can easily show for the other cases that they also satisfy $\min\{\abs{h}, \abs{c}\}/16 \le \Rob\nomtDR(h, c) \le 1 + \min\{\abs{h}, \abs{c}\}$.
.

\iffollowingorders
{\bf Case $m\ge 1$, $u \ge 1$ and $w \ge 1$.}
\else
\paragraph{Case \bm{$m\ge 1$}, \bm{$u \ge 1$} and \bm{$w \ge 1$}.}
\fi
We can ignore the instances that are drawn from the error region.
If $x\in\dis{h, c}$, then 
these instances do not increase the robustness of $h$.
By (\ref{eq:dis-region:count}) such instances are drawn with probability $\mu = (2^{w} + 2^u - 2)\cdot 2^{-m-u-w}$.
%

\iffollowingorders
{\bf Lower Bound.}
\else
\paragraph{Lower Bound.}
\fi
We examine the case where $\abs{\concept} \le \abs{\hconcept}$.
The case $\abs{\concept} > \abs{\hconcept}$ can be handled symmetrically.
First, $\abs{\concept} \le \abs{\hconcept} 
\Rightarrow 
u \le w$.
We now look at the following falsifying truth assignments 
in order to obtain a lower bound.
Let $x$ falsify $1\le j \le m$ variables that mutually appear between $h$ and $c$.
Further, let $x$ falsify $\zeta\ge 1$ more variables among the $u$ that appear only in $c$
and $\xi\ge 0$ more variables among the $w$ that appear only in $h$. 
Then, we can perturb $x$ into an $x'\in\dis{h, c}$ by changing $\sigma = j + \min\{\zeta, \xi\}$ coordinates.
The contribution to the overall robustness is then,  
\begin{displaymath}
2^{-m-w-u}\sum_{j=1}^m\sum_{\zeta=1}^u\sum_{\xi=0}^{r} \binom{m}{j}\binom{u}{\zeta}\binom{w}{\xi}(j + \min\{\zeta, \xi\})\,.
\end{displaymath}
Letting the above quantity be $Q$ we have, 
\begin{align*}
Q 
&= \phantom{+} 2^{-m-w-u}\sum_{j=1}^m j\binom{m}{j}\sum_{\zeta=1}^u \binom{u}{\zeta} \sum_{\xi=0}^{w} \binom{w}{\xi}
\\
&\phantom{=} + 2^{-m-w-u} \sum_{j=1}^m \binom{m}{j} \sum_{\zeta=1}^u \zeta \binom{u}{\zeta} \sum_{\xi=\zeta}^{w} \binom{w}{\xi} \\ 
&\phantom{=} + 2^{-m-w-u} \sum_{j=1}^m \binom{m}{j} \sum_{\zeta=1}^u \binom{u}{\zeta} \sum_{\xi=0}^{\zeta-1} \xi\binom{w}{\xi}\,. 
\end{align*}
where, by dropping the last term and 
applying Lemma 
\ref{lem:important:lower-bound} 
we obtain, 
$Q 
\ge 
2^{-m-u}(2^{u}-1)m2^{m-1} 
+ \frac{u}{8}\cdot 2^{u+w}\cdot 2^{-m-w-u} \sum_{j=1}^m \binom{m}{j} 
$.
Thus, 
$Q 
\ge \frac{m}{2}(1 - 2^{-u}) + \frac{u}{8}(1 - 2^{-m}) \ge \min\{\abs{h}, \abs{c}\}/16 $. 

\iffollowingorders
{\bf Upper Bound.}
\else
\paragraph{Upper Bound.}
\fi
First, if $x\in\dis{h, c}$, then we need to modify $\sigma = 0$ coordinates in $x$ so that the perturbed instance $x'$ is in the error region.
Second, 
if $x$ is a satisfying truth assignment for both $h$ and $c$, then since $h$ and $c$ differ in at least one variable, making that variable in $x$ equal to $0$ will result in an $x'\in\dis{h, c}$. Therefore,
we need to modify $\sigma = 1$ coordinates in $x$
in this case so that $x'\in\dis{h, c}$.
Finally, if $x$ is a falsifying truth assignment for both $h$ and $c$, then the minimum perturbation 
is obtained by flipping $\sigma \le (1+ \min\{\abs{\hconcept}, \abs{\concept}\})$ 0's to 1's. Therefore, we need to perturb $\sigma \le 1 + \min\{\abs{\hconcept}, \abs{\concept}\}$ coordinates in $x$ in every case so that for the perturbed instance $x'$ it holds $x'\in\dis{h, c}$.
\end{proof}
\fi


\iffollowingorders
\else
\subsection{Prediction Change Risk and Robustness}

%
%

\begin{restatable}[Prediction change risk]{theorem}{ThmPCRisk}\label{thm:risk:mon-conj:prediction-change}
Consider the Problem Setup \ref{setup:mon-conj}.
Then, 
$\Risk\nomPC(h) = 0$ when $r = 0$,
while when $r \ge 1$ we have, 
\begin{displaymath}
\Risk\nomPC(h) = 2^{-\abs{h}}\cdot\sum_{i=0}^r\binom{\abs{h}}{i}\,.
\end{displaymath}
\end{restatable}
\iffollowingorders
\else
\begin{proof}
With a budget of $r = 0$ we can not change the evaluation of $h$ at any instance, so the prediction change adversarial risk is $0$.

Now consider the case where we have a budget of $1\le r < \abs{h}$.
A satisfying truth assignment $x$ (which arises with probability $2^{-\abs{h}}$) requires a budget of only $r = 1$
in order to become falsifying.
%
For a falsifying truth assignment $x$ that has $i$ 0's among 
the $\abs{h}$ variables that appear in $h$ we need a budget of $i$. 
Therefore, for a budget of $1\le r < \abs{h}$
we can violate $\sum_{i=1}^r\binom{\abs{h}}{i}2^{n-\abs{h}}$ truth assignments
that were originally falsifying.

Of course if we have a budget of $r\ge \abs{h}$, then we can change any $x$ to an $x'$
such that $h(x) \neq h(x')$ and therefore the overall risk is going to be $1$.
In other words, for the interesting cases where $1\le r < \abs{h}$ we have 
$\Risk\nomPC(h) = 2^{-\abs{h}} + 2^{-n}\cdot\sum_{i=1}^{r}\binom{\abs{h}}{i}2^{n-\abs{h}} = 2^{-\abs{h}}\sum_{i=0}^{r}\binom{\abs{h}}{i}$. This last formula is also consistent with budget $r \ge \abs{h}$ as then $\Risk\nomPC(h) = 1$.
\end{proof}
\fi
\fi

%
%
\iffollowingorders
\begin{restatable}{theorem}{ThmPCRobustness}\label{thm:robustness:mon-conj:prediction-change}
Consider the Problem Setup \ref{setup:mon-conj}.
Then, $\Rob\nomtPC(h) = \abs{h}/2 + 2^{-\abs{h}}$.
\end{restatable}
\else
\begin{restatable}[Prediction change robustness]{theorem}{ThmPCRobustness}\label{thm:robustness:mon-conj:prediction-change}
Consider the Problem Setup \ref{setup:mon-conj}.
Then,
\begin{displaymath}
\Rob\nomtPC(h) = \frac{1}{2}\cdot\abs{h} + 2^{-\abs{h}}\,.
\end{displaymath}
\end{restatable}
\fi
\iffollowingorders
\else
\begin{proof}
When $x$ is a satisfying truth assignment for \hconcept then it suffices to turn a single 1 into a 0 for one of the variables that appear in \hconcept so that the resulting $x'$ is such so that $h(x) \neq h(x')$.
%
When $x$ is a falsifying truth assignment for \hconcept with $i$ out of the $\abs{h}$ variables that appear in \hconcept being 1,
then we need to flip $\sigma = \abs{h} - i$ bits in $x$ for the prediction to change.
Therefore, 
$
\Rob\nomtPC(h)
= 2^{-n}\cdot 1\cdot 2^{n-\abs{h}} + 2^{-n}\cdot\sum_{i=0}^{\abs{h}-1} (\abs{h} - i)\cdot\binom{\abs{h}}{i}\cdot 2^{n-\abs{h}} 
= 2^{-\abs{h}} +  2^{-\abs{h}}\cdot\sum_{j=1}^{\abs{h}} j\cdot\binom{\abs{h}}{j} 
= 2^{-\abs{h}} + \abs{h}/2$.  
\end{proof}
\fi


\iffollowingorders
\else
\subsection{Corrupted Instance Risk and Robustness}

%
%
In the theorem below the important message is that driving the adversarial risk to 1 based on the corrupted instance definition requires budget $\Theta\left(\abs{h}\right)$ contrasting the budget of size $\Theta\left(\min\set{\abs{h}, \abs{c}}\right)$ that is required in Theorem \ref{thm:dis:risk} in order to drive the adversarial risk to 1 based on the error region definition. The other cases in the statement below refer to intermediate values of budget and explain how we arrive at this conclusion.
\begin{restatable}[Corrupted instance risk]{theorem}{ThmCIRisk}\label{thm:risk:original-truth}
Consider the Problem Setup \ref{setup:mon-conj}.
Then, for a budget of size $r$ we have, 
\begin{itemize}
\item if $r = 0$, then $\Risk\nomCI(h, c) = \mu = (2^w + 2^u - 2)\cdot 2^{-m-u-w}$\,,
\item if $1\le r < w$, 
then 
$\Risk\nomCI(h, c) = 2^{-\abs{c}} 
+ 2^{-\abs{h}}
\left(\sum_{j=0}^{r}\binom{\abs{h}}{j} 
- 2^{-u}\sum_{\xi=0}^r\binom{w}{\xi}\right)$\,,
\item if $w \le r < \abs{h}$, 
then 
$\Risk\nomCI(h,c) = 2^{-\abs{h}}\cdot\sum_{j=0}^{r}\binom{\abs{h}}{j}$\,,
\item if $\abs{h} \le r$, then 
$\Risk\nomCI(h, c) = 1$\,.
\end{itemize}
\end{restatable}
\iffollowingorders
\else
\begin{proof}
When $x\in\dis{h, c}$, 
all these instances contribute to the adversarial risk, for any budget value (including $r = 0$), since they satisfy $h(x) = h(x') \neq c(x)$. The probability of obtaining such an $x$ is, by (\ref{eq:dis-region:count}), $(2^w + 2^u - 2)\cdot 2^{-m-u-w}$.

\medskip

When $x$ is a satisfying truth assignment for both $h$ and $c$, then 
we need to flip $\sigma = 1$ bits. 
Therefore with a budget of $1$, when either $h$ or $c$ (or both) are satisfied, 
then these truth assignments contribute to the risk an amount of
$2^{-m-u} + 2^{-m-w} - 2^{-m-u-w}$.

\medskip

When $x$ is a falsifying truth assignment for both $h$ and $c$ we have the following two cases.
\begin{itemize}
\item 
When $x$ falsifies $1\le j \le m$ mutual and $0 \le \xi \le w$ wrong variables,
we need to perturb $\sigma = j+\xi$ bits so that
$h(x') \neq c(x)$.
Therefore with a budget of $r$ we can increase the risk by an amount of 
\begin{displaymath}
2^{-m-w}\sum_{j=1}^{r}\left(\binom{m+w}{j} - \binom{w}{j}\right)\,.
\end{displaymath}

\item 
When no mutual variable is falsified 
(or there are no mutual variables between $h$ and $c$),
but $x$ falsifies 
$1\le \zeta\le u$ undiscovered and $1\le \xi \le w$ wrong variables,
then we need to perturb $\sigma = \xi$ bits.
Hence, with a budget of $r$ we can increase the risk by an amount of 
$2^{-m-u-w}(2^u-1)\sum_{\xi=1}^r\binom{w}{\xi}$.
\end{itemize}

We now add everything up. 
\end{proof}
\fi
\fi

%
%
\iffollowingorders
\begin{restatable}{theorem}{ThmCIRobustness}\label{thm:robustness:mon-conj:original-truth}
Consider the Problem Setup \ref{setup:mon-conj}. Then, $\abs{h}/4 < \Rob\nomtCI(h, c) < \abs{h} + 1/2$.
\end{restatable}
\else
\begin{restatable}[Corrupted instance robustness]{theorem}{ThmCIRobustness}\label{thm:robustness:mon-conj:original-truth}
Consider the Problem Setup \ref{setup:mon-conj}. Then, 
\begin{displaymath}
\frac{1}{4}\cdot\abs{h} < \Rob\nomtCI(h, c) < \abs{h} + \frac{1}{2}\,.
\end{displaymath}
\end{restatable}
\fi
\iffollowingorders
\else
\begin{proof}
First note that instances in the error region contribute precisely $0$ to the overall robustness, regardless if we are looking for a lower bound or an upper bound on the robustness.

For the lower bound 
consider 
truth assignments that are falsifying both $h$ and $c$
such that the following variables are falsified: 
$1\le j \le m$ mutual variables, 
$0\le \zeta \le u$ undiscovered variables, 
and
$0\le \xi \le w$ wrong variables. 
So that we can achieve $h(x') \neq c(x)$, 
we need an $x'$ such that $h(x') = 1$.
But then this means that we need to flip 
$\sigma = j + \xi$ bits in $x$ in this case. 
As a result, the contribution of these truth assignments to the overall robustness of $h$ is,
\begin{eqnarray}
Q_1 &=& 2^{-m-w-u}\sum_{j=1}^m \binom{m}{j}\sum_{\zeta = 0}^u \binom{u}{\zeta} \sum_{\xi = 0}^w \binom{w}{\xi}\cdot(j+\xi) \nonumber\\
 &=& \frac{m}{2} + \frac{w}{2}\cdot(1 - 2^{-m}) 
 \ge \frac{m}{2} + \frac{w}{4}
 > \frac{m+w}{4} = \frac{\abs{\hconcept}}{4}\,. \nonumber
\end{eqnarray}

For the upper bound, for a truth assignment $x$ that is falsifying both $h$ and $c$ we need to change no more than $\abs{h}$ bits in $x$ so that for the perturbed instance $x'$ it holds $h(x') = 1 \neq 0 = h(x) = c(x)$. As the probability of obtaining such a truth assignment $x$ is strictly less than $1$, then the contribution to the overall robustness due to such truth assignments is strictly less than $\abs{h}$.
On the other hand, for satisfying truth assignments of both $h$ and $c$ we have $\sigma = 1$ bits that need to change in $x$ and therefore the contribution to the overall robustness is $2^{-m-u-w} \le 1/2$.
\end{proof}
\fi


\subsection{Experiments for the Expected Values of Adversarial Robustness}\label{sec:experiments:conjunctions}

In this part, we complement the theorems that we presented earlier with experiments. This way we are able to examine how some popular algorithms behave under attack, and we explore the extent to which the generated solutions of such algorithms exhibit differences in their (adversarial) robustness on average against various target functions drawn from the class of monotone conjunctions.

\iffollowingorders
\else
\paragraph{Overview of algorithms.}
\fi
The first algorithm is the standard Occam algorithm that starts from the full conjunction and eliminates variables from the hypothesis that contradict the positive examples received; this algorithm is known as \finds in \citep{Book::Mitchell:MachineLearning} 
\iffollowingorders
\else
because it maintains the most \emph{specific} hypothesis in the \emph{version space} \citep{Mitchell::VersionSpaces} induced by the training data, 
\fi
but has appeared without a name earlier by Valiant in \citep{Valiant:PAC} and its roots are at least as old as in  \citep{Book:Thinking}.
The second algorithm is the \swapping from the framework of evolvability \citep{Valiant::Evolvability}. This algorithm searches for an $\eps$-optimal solution among monotone conjunctions that have at most $\lceil\lg(3/(2\eps))\rceil$ variables in their representation using a local search method where hypotheses in the neighborhood are obtained by swapping in and out some variable(s) from the current hypothesis; 
\iffollowingorders
\else
in particular 
\fi
we follow the analysis that was used in \citep{Diochnos::saga2009} and is a special case of the analysis used in~\citep{Diochnos::ALT2016}.

\iffollowingorders
\else
\paragraph{Experimental setup.}
\fi
In each experiment, we first learn hypotheses by using the algorithms under \UUn against different target sizes. 
For both algorithms, during the learning process, we use $\eps = 0.01$ and $\delta = 0.05$ for the learning parameters. 
We then examine the robustness of the generated hypotheses by drawing examples again from the uniform distribution \UUn as this is the main theme of this paper. 
In particular, we test against the 30 target sizes from the set 
$\{1, 2, \ldots, 24, 25, 30, 50, 75, 99, 100\}$.
For each such target size, we plot the average value, over 500 runs, of the robustness of the learned hypothesis that we obtain. In each run, we repeat the learning process 
using 
a random target of the particular size 
as well as a fresh training sample 
and subsequently estimate the robustness of the learned hypothesis
by drawing another $10,000$ examples from \UUn that we violate (depending on the definition). The dimension of the instances is $n = 100$.

\begin{figure}[ht]
    \centering
    {
    \iffollowingorders
    \includegraphics[width=0.7\columnwidth]{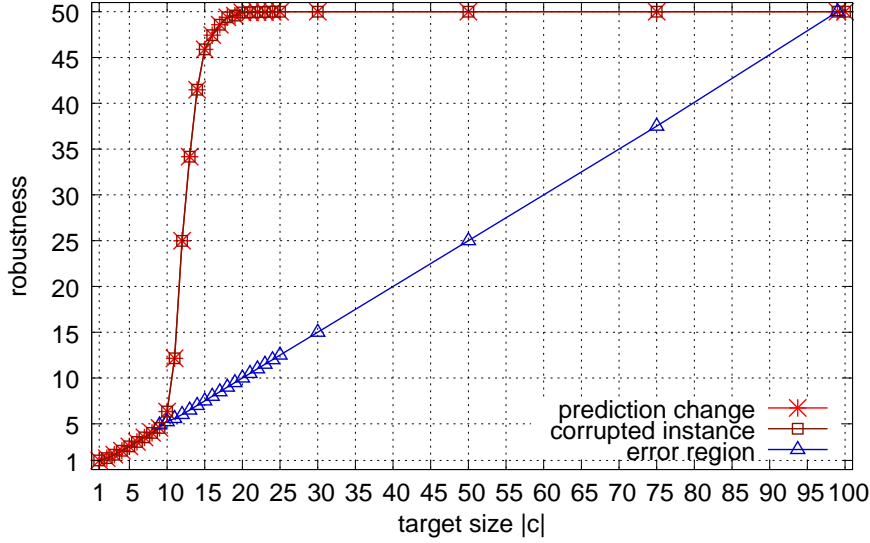}
    \else
    \includegraphics[width=0.7\textwidth]{PAC_robustness.eps}
    \fi
    }
    \caption{Experimental comparison of the different robustness measures.
    The values for PC and CI almost coincide and they can hardly be distinguished.
    The value for ER robustness is completely different compared to the other two.
    Note that ER robustness is $\infty$ 
    when the target size $\abs{c}$ is in $\{1, \ldots, 8\}\cup\{100\}$ 
    and for this reason only the points between 9 and 99 are plotted.
    When $\abs{c} \ge 20$, almost always the learned hypothesis is the initialized full conjunction.
    The reason is that positive examples are very rare and our training set contains none.
    As a result no variable is eliminated from the initialized hypothesis $h$ (full conjunction).
    Hence, when $\abs{c} \ge 20$ we see that PC and CI robustness is about 
    ${\max}\{\abs{h}, \abs{c}\}/2 = {\abs{h}}/2$,
    whereas ER is roughly ${\min}\{\abs{h}, \abs{c}\}/2 = {\abs{c}}/2$.
    }
    \label{fig:robustness:mon-conj:pac}
\end{figure}

\iffollowingorders
Figure \ref{fig:robustness:mon-conj:pac} presents the values of the three robustness measures for the case of \finds. 
In the full version of the paper we provide more details on the algorithms and more information regarding our experiments. 
The message is that the adversarial robustness that is based on the definitions of \emph{prediction change} and \emph{corrupted instance} is more or less the same, whereas the adversarial robustness based on the \emph{error region} definition may obtain wildly different values compared to the other two. 
\else
\subsubsection{The Algorithm \finds}
Algorithm \finds is shown in Algorithm \ref{algo:pac}.
The initial hypothesis is the full monotone conjunction and as positive examples are received during training, the learner drops from the hypothesis those variables that are falsified in the positive examples.
Following the Problem Setup \ref{setup:mon-conj}, 
the concept class \Ccn contains $\abs{\Ccn} = 2^n - 1$ monotone conjunctions since the empty conjunction is not part of the class. Similarly, the hypothesis class is $\hypothesisc = \Ccn$.
So, in particular for our experiments we have $\abs{\hypothesisc} = 2^{100} - 1$ since $n=100$.
Therefore, we use a sample of size 
$m = \lceil\frac{1}{\eps}\ln\left(\abs{\hypothesisc}/\delta\right)\rceil = 7232$ which is enough (see, e.g., \citep{OccamsRazor}) for learning any finite concept class of size $\abs{\hypothesisc}$ up to risk $\eps$, with probability at least $1-\delta$, under any distribution and in particular under the uniform distribution \UUn as in our experiments. 

\begin{algorithm}[ht]
{
\footnotesize

\KwIn{
dimension $n\in\NN^*$, 
$\confidence\in(0, 1)$, 
$\eps\in (0, 1)$}
\KwOut{a hypothesis $\hconcept\in\Cc_n$}
$\hconcept \assign \wedge_{i=1}^n x_i$\;
Draw $m = \left\lceil\frac{1}{\eps}\cdot\ln\left(\abs{\hypothesisc}/\confidence\right)\right\rceil$ examples\;
\For{$j \assign 1$ up to $m$}{
	\lIf{$j$-th example is positive}{
		remove from \hconcept all the variables that are falsified in the 
		example
	}
}
\Return \hconcept;
\caption{An Occam algorithm for efficiently PAC learning monotone conjunctions}\label{algo:pac}
}
\end{algorithm}

As we proved in Theorems \ref{thm:robustness:mon-conj:prediction-change} and \ref{thm:robustness:mon-conj:original-truth} the robustness of the learned hypothesis is \emph{proportional to the size of the hypothesis} regardless of the size of the target when using the definitions of \emph{prediction change} and \emph{corrupted instance}
(and in fact, experimentally the robustness in both cases is about $\abs{h}/2$).
However, the robustness of the hypothesis using the \emph{error region} approach is \emph{proportional to the minimum between the size of the hypothesis and the size of the target}. 
Now note that the way \finds constructs solutions, it holds that $\abs{c} = \min\set{\abs{h}, \abs{c}}$. 
As such, when the target is sufficiently large (e.g., $\abs{c} \ge 20$), then the learned hypothesis using \finds is almost always the initial full conjunction of size 100 and therefore the robustness based on the \emph{error region} approach is proportional to $\abs{c}$ and not proportional to 100 (which is the size of $h$) as it is the case of the robustness that is based on the \emph{prediction change} and \emph{corrupted instance} definitions. 
In other words, the robustness obtained by the \emph{error region} approach is significantly different from the robustness obtained by the \emph{prediction change} and \emph{corrupted instance} approaches for a significant range of target functions. Figure \ref{fig:robustness:mon-conj:pac} presents the values of the three robustness measures when using the \finds algorithm to form a hypothesis
in our setup ($\eps = 0.01$, etc.).
Then, Figure \ref{fig:mon-conj:pac:size} presents the average size of the learned hypotheses and Figure \ref{fig:mon-conj:pac:error} presents the average error of the hypotheses learned.
\fi

\iffollowingorders
\else
\begin{figure}[ht]
    \centering
    {
    \iffollowingorders
    \includegraphics[width=0.7\columnwidth]{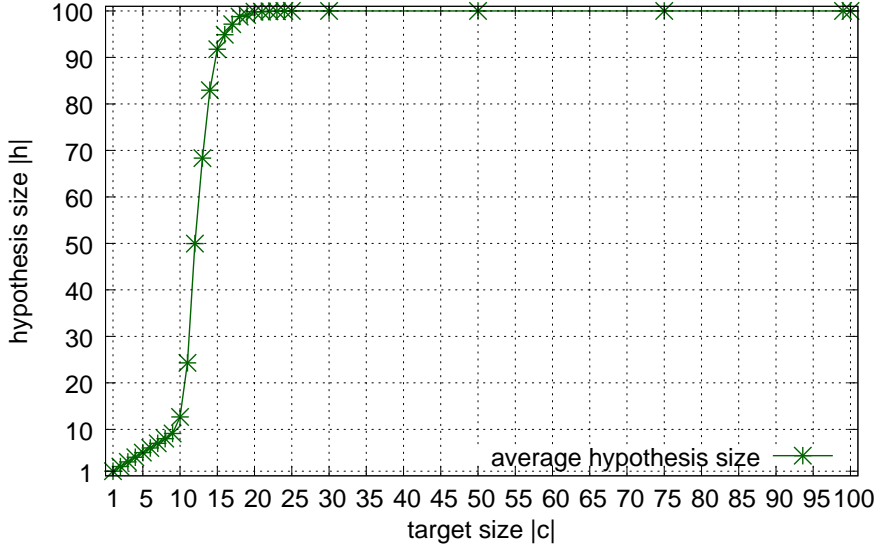}
    \else
    \includegraphics[width=0.7\textwidth]{PAC_size_avg.eps}
    \fi
    }
    \caption{Average size, over 500 runs, for the computed hypothesis using \finds for learning monotone conjunctions with a sample of size $7,232$ under the uniform distribution \UUn.}
    \label{fig:mon-conj:pac:size}
\end{figure}

\begin{figure}[ht]
    \centering
    {
    \iffollowingorders
    \includegraphics[width=0.7\columnwidth]{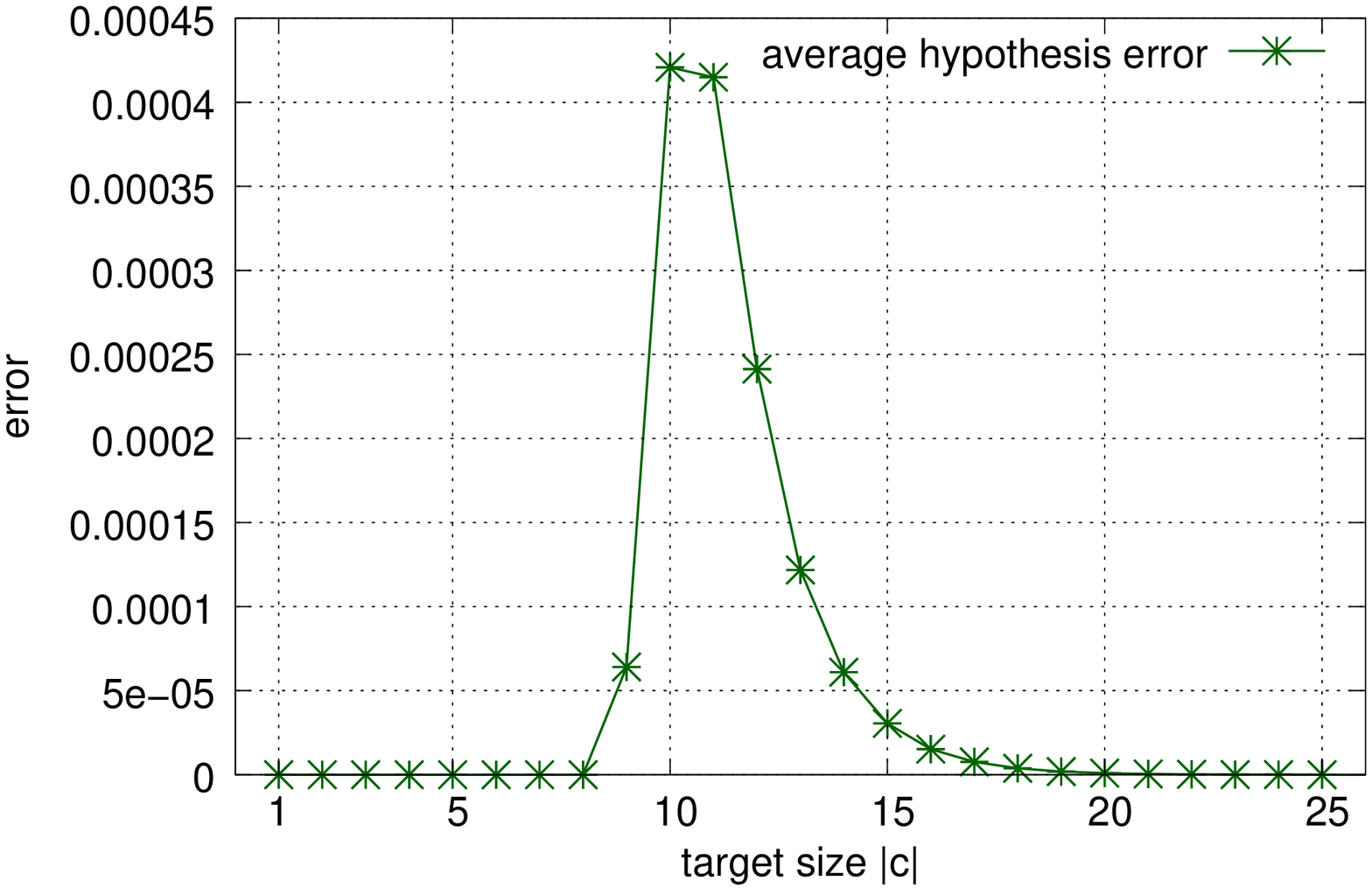}
    \else
    \includegraphics[width=0.7\textwidth]{PAC_error_avg.eps}
    \fi
    }
    \caption{Average error, over 500 runs, of the computed hypotheses using \finds for learning monotone conjunctions with a sample of size $7,243$ under the uniform distribution \UUn.
    For target sizes $\abs{c}\in \{1, \ldots, 8\}\cup\{100\}$, the learner always identifies precisely the target and thus the error is precisely 0 in that regime.
    (For the case $\abs{c} = 100$ the learner is lucky, as the initial guess is the target itself and thus the hypothesis never changes as examples are being presented to the learner.) 
    For target sizes $\abs{c}$ between 14 and 99 the average error is nonzero 
    but less than $10^{-4}$ and decays as the size of the target increases.
    For clarity we only plot target sizes between 1 and 25.}
    \label{fig:mon-conj:pac:error}
\end{figure}

\subsubsection{The \swapping}
The \swapping appeared in Valiant's framework of evolvability \citep{Valiant::Evolvability}. We  give a brief description of this algorithm without going deeply in our discussions regarding fascinating details of the framework of evolvability. 
Loosely speaking, in evolvability the idea is to develop an evolutionary mechanism that allows near-optimal hypotheses to form by letting such hypotheses interact with the environment. Here we describe the evolutionary mechanism as if it is deployed by some learner who knows how the evolutionary mechanism operates on hypotheses (organisms), how these hypotheses interact with the environment, and  how this interaction affects the hypothesis formation from one iteration (generation) to another.

First of all, 
the algorithm 
is a \emph{local-search} method. 
At any given time 
the algorithm is maintaining a hypothesis that is a monotone conjunction of some of the Boolean variables. 
For any given hypothesis that the learner has formed, a \emph{mutator} function 
\begin{inparaenum}[(i)]
\item defines a \emph{neighborhood} $N(h)$\footnote{We will simply write $N$ instead of $N(h)$, since the neighborhood always depends on the current hypothesis $h$.} with hypotheses that can potentially replace the current hypothesis of the learner, 
\item \emph{scores} the candidates in the neighborhood based on their \emph{predictive performance} on samples of certain size, and
\item uses a \emph{decision rule} to determine which hypothesis among the ones in the neighborhood will be selected as the hypothesis that the learner will have during the next iteration (generation).
\end{inparaenum}

\iffollowingorders
\begin{asparadesc}
\item [Neighborhood.] 
\else
\paragraph{Neighborhood.}
\fi
The neighborhood $N$, can, in general, be decomposed into $N = N^+\cup N^-\cup N^{\pm}\cup\set{h}$.
The neighborhood $N^+$ contains the hypotheses that have one more variable compared to the current hypothesis $h$, the neighborhood $N^-$ contains the hypotheses that have one less variable compared to the current hypothesis $h$, and the neighborhood $N^{\pm}$ contains the hypotheses that are obtained by \emph{swapping} one of the variables that appear in $h$ with one of the variables that do not yet appear in $h$.
As an example, let $h = x_1 \wedge x_2$ and $n = 3$. Then, $N^- = \set{x_1, x_2}$, $N^+ = \set{x_1\wedge x_2 \wedge x_3}$, and $N^{\pm} = \set{x_1 \wedge x_3, x_2\wedge x_3}$.
As we can see, the current hypothesis $h$ is always in the neighborhood, so the learner can always retain its current guess for the next iteration.\footnote{Additions and deletions can be seen as swaps with the constant $1$. This swapping nature of the algorithm justifies its name that was given in \citep{Diochnos::saga2009}.}

\iffollowingorders
\item [Weights.]
\else
\paragraph{Weights.}
\fi
Each hypothesis in $N^+\cup N^-\cup \set{h}$ is assigned the same weight so that all these weights add up to $1/2$. Each hypothesis in $N^\pm$ is assigned the same weight so that the weights of all these hypotheses in $N^\pm$ also add up to $1/2$.

\iffollowingorders
\item [Scoring.] 
\else
\paragraph{Scoring.}
\fi
The predictive power of each hypothesis in 
$N$ is approximated, with high probability, within $\upepsilon_s$ of its true value, by testing its \emph{predictive performance} on a sufficiently large sample.

\iffollowingorders
\item [Partitioning.] 
\else
\paragraph{Partitioning.}
\fi
The hypotheses in $N$ are partitioned into three sets based on their \emph{predictive performance}. For a real constant $t$, called \emph{tolerance}, and a base performance value $\upnu_h$ for the current hypothesis $h$, the hypotheses that exhibit performance strictly larger than $\upnu_h + t$ form the \emph{beneficial} group $Bene$, the hypotheses that exhibit performance within $\upnu_h \pm t$ form the \emph{neutral} set $Neut$, and finally the hypotheses that exhibit performance strictly less than $\upnu_h - t$ form a \emph{deleterious} set.

\iffollowingorders
\item [Decision rule for mutator.] 
\else
\paragraph{Decision rule for mutator.}
\fi
If the set $Bene$ is non-empty, then a hypothesis is selected from this set, otherwise a hypothesis from the set $Neut$ is selected. Note that $Neut$ is always nonempty as the current hypothesis is always there. When a hypothesis has to be selected among many from a set (whether $Bene$ or $Neut$), it is selected with probability proportional to its weight compared to the total weight of the hypotheses in the set.

\iffollowingorders
\item [Hypothesis class \hypothesisc.]
\else
\paragraph{Hypothesis class \hypothesisc.}
\fi
For a threshold $q = \left\lceil\log_2(3/(2\eps))\right\rceil$ it can be shown \citep{Diochnos::saga2009} that, for any target monotone conjunction from our concept class \Ccn in the Problem Setup \ref{setup:mon-conj}, among the monotone conjunctions that have up to $q$ variables, there is at least one such conjunction that has risk at most $\eps$. In particular, the \swapping will 
\begin{inparaenum}[(i)]
\item identify precisely targets with up to $q$ variables (i.e., generate a hypothesis that achieves risk $0$), or
\item form a hypothesis with $q$ variables that has risk at most $\eps$ for targets that have size larger than $q$.
\end{inparaenum}
Due to this property of the algorithm, a typical hypothesis class is $\hypothesisc = \Ccn^{\le q}$ that contains the monotone conjunctions that have $0$ up to $q$ variables. Therefore, we formally deviate from the hypothesis class $\hypothesisc = \Ccn$ found in our Problem Setup \ref{setup:mon-conj} - even though the algorithm can still run with this hypothesis class and moreover in the end the algorithm will return a hypothesis that satisfies $h\in\Ccn$ as in the Problem Setup \ref{setup:mon-conj}. This deviation on the hypothesis class is done on one hand for simplicity because it is easier to explain the evolutionary mechanism, and on the other hand because such a selection of a hypothesis class technically gives an algorithm in the \emph{non-realizable} case of learning - which might be interesting in its own right.
For more information on these details we refer the reader to \citep{Diochnos::saga2009} and \citep{Diochnos::ALT2016}.
\iffollowingorders
\end{asparadesc}
\else\fi

The mutator function for the \swapping is shown in Algorithm \ref{algo:swapping:uniform}.
\begin{algorithm}[ht]
{
\footnotesize
\SetKwData{NeighborhoodSize}{$\upsigma$}
\SetKwData{tol}{$t$}
\SetKwData{Bene}{$Bene$}
\SetKwData{Neutral}{$Neutral$}
\SetKwData{BaseCorr}{$\upnu_h$}
\SetKwFunction{ComputePerformance}{Perf}
\SetKwFunction{GetPerformance}{GetPerformance}
\SetKwFunction{SetPerformance}{SetPerformance}
\SetKwFunction{Mutator}{Mutator}
\SetKwFunction{NeighborsAddOneLiteral}{NborsAddOne}
\SetKwFunction{NeighborsRemoveOneLiteral}{NborsRemoveOne}
\SetKwFunction{NeighborsPureMutate}{NborsPureMutate}
\SetKwFunction{SetWeight}{SetWeight}
\SetKwFunction{RandomSelect}{Select}
\SetKwFunction{SetTolerance}{SetTolerance}
\SetKwFunction{ComputeQ}{Compute-q}
\SetKwFunction{ComputeMinNonZeroA}{ComputeMinNonZeroA}

\KwIn{dimension $n$, 
$\confidence\in(0, 1)$, 
$\eps\in (0, 1)$,
$\hconcept\in\hypothesisc$}
\KwOut{a new hypothesis $\hconcept'$}
$q \assign \left\lceil\log_2(3/(2\eps))\right\rceil$\;

\leIf{$\abs{\hconcept} > 0$}{Generate $N^-$}{$N^-\assign\emptyset$}
\leIf{$\abs{\hconcept} < q$}{Generate $N^+$}{$N^+\assign\emptyset$}
\leIf{$\abs{\hconcept} \leq q$}{Generate $N^{\pm}$}{$N^{\pm}\assign\emptyset$}
$\Bene \assign \emptyset$; $\Neutral \assign \{\hconcept\}$\;
$t \assign 2^{-2q}$; \ 
$\upepsilon_s \assign 2^{-2q}$; \ 
$\confidence_s \assign \confidence / (6q^2n)$;

\SetWeight{$\hconcept$, \hconcept, $N^-$, $N^+$, $N^{\pm}$}; 
\BaseCorr \assign \ComputePerformance{\hconcept, $\upepsilon_s$, $\confidence_s$}\; 
\For{$x \in N^+, N^-, N^{\pm}$}{
  \SetWeight{$x$, \hconcept, $N^-$, $N^+$, $N^{\pm}$}; 
  $\upnu_x$ \assign \ComputePerformance{$x$, $\upepsilon_s$, $\confidence_s$}\;
  \lIf{$\upnu_x > \BaseCorr + \tol$}{
    \Bene \assign \Bene $\cup \ \{x\}$ 
  }
  \lElseIf{$\upnu_x \geq \BaseCorr - \tol$}{
    \Neutral \assign \Neutral $\cup \ \{x\}$ 
  }
}
\leIf{\Bene $\neq \ \emptyset$}{\Return \RandomSelect{\Bene}}{\Return \RandomSelect{\Neutral}}
\caption{Mutator function for uniform distribution}\label{algo:swapping:uniform}
}
\end{algorithm}
The function \texttt{SetWeight} assigns weights to the hypotheses as explained in the paragraph about assigning weights above. The function \texttt{Select} returns a hypothesis from the set in its argument as explained in the paragraph above regarding the decision rule for the mutator. The function \texttt{Perf} computes empirically the predictive performance of the hypotheses within accuracy $\epsilon_s$ with probability at least $1-\confidence_s$.

As mentioned above, for the solutions obtained by the \swapping it holds that $\abs{h} = \min\set{\abs{h}, \abs{c}}$. Thus, while we can observe some differences for a small range of target functions regarding the robustness of the generated hypotheses based on the various definitions, these differences are smaller compared to what we obtain with the \finds algorithm. 
Figure \ref{fig:robustness:mon-conj:evo:zoom-out} presents the values of the three robustness measures 
when using the swapping algorithm from the framework of evolvability in order to form a hypothesis.
The robustness measures based on the \emph{prediction change} and \emph{corrupted instance} definitions behave very similarly against the solutions obtained by the \swapping, just like they did for the solutions obtained by the \finds algorithm.
\begin{figure}[H]
    \centering
    {
    \iffollowingorders
    \includegraphics[width=0.7\columnwidth]{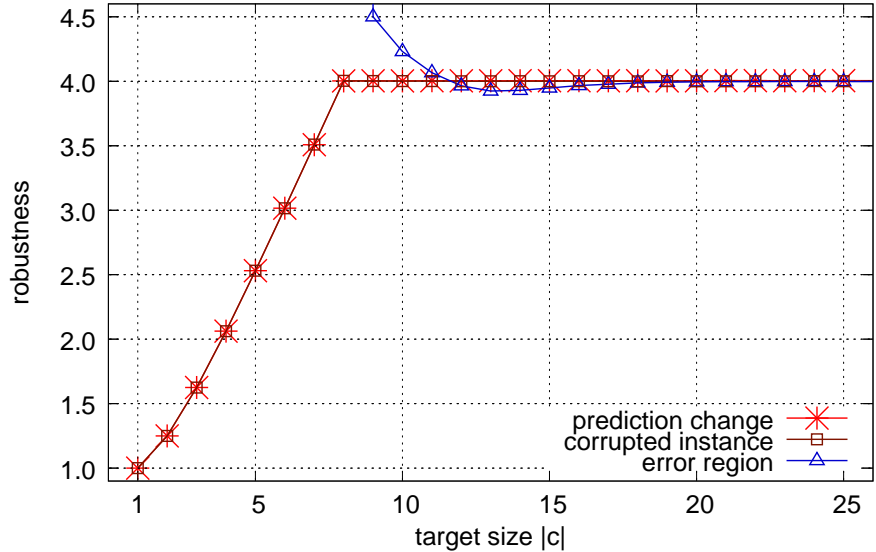}
    \else
    \includegraphics[width=0.7\textwidth]{EVO_robustness_zoom_out.eps}
    \fi
    }
    \caption{Experimental comparison of the different robustness measures regarding the hypotheses that are obtained by the \swapping.
    The values for all three measures almost coincide when the target has size $\abs{c} \ge 20$ and for clarity we only plot the values until $\abs{c} = 25$.
    Note that ER robustness is $\infty$ 
    when the target size $\abs{c}$ is in $\{1, \ldots, 8\}\cup\{100\}$ 
    and for this reason these points are not plotted (similarly to Figure \ref{fig:robustness:mon-conj:pac}).
    Further, when $9 \le \abs{c} \le 19$, we can observe that ER robustness behaves slightly differently
    compared to PC and CI and eventually in the regime $20 \le \abs{c} \le 99$ it is much closer
    to CI with PC being consistently slightly larger than both of them 
    (and this is explained by the fact that truth assignments that belong to the disagreement region still need to be perturbed).
    Finally, our earlier bounds predict that the values will be roughly the same
    as this time $\min\{\abs{h}, \abs{c}\} = \abs{h}$
    contrasting the observation from Figure \ref{fig:robustness:mon-conj:pac}.}
    \label{fig:robustness:mon-conj:evo:zoom-out}
\end{figure}
We see that the robustness due to the \emph{error region} is different for some targets of certain size compared to the robustness values obtained when using the \emph{prediction change} and the \emph{corrupted instance} definitions, but this time the differences are smaller compared to what we observe for the solutions obtained by the \finds algorithm (and of course excluding the cases where the error region robustness is infinite). 
However this is more or less expected since, on one hand for the learned solution it is always true that $\min\{\abs{h}, \abs{c}\}$ coincides with $\abs{h}$ which is the factor governing the robustness under the \emph{prediction change} and \emph{corrupted instance} definitions, regardless of the size of the target $\abs{c}$, 
and on the other hand, again due to the fact that $\abs{h} = \min\{\abs{h}, \abs{c}\}$, the robustness of the generated solutions have significantly smaller values compared to the robustness of the generated solutions obtained by \finds.

\fi

\section{Inherent Bounds on Risk and Robustness for the Uniform Distribution} \label{sec:uniform}
After showing the different behavior of various definitions of adversarial risk and robustness through a study of monotone conjunctions under the uniform distribution, in this section, we state 
\iffollowingorders
\else
and prove
\fi
our main theorems about \emph{inherent barriers} against achieving \emph{error region} adversarial risk and robustness of \emph{arbitrary} learning problems whose instances are distributed uniformly over the $n$-dimension hypercube $\bits^n$. 
\iffollowingorders
The proofs of the theorems below are available in the full version of the paper.
\else
\fi 

We first define a useful notation for the size of the (partial) Hamming balls and state three lemmas and two corollaries based on the notation.

\begin{definition}
For every $n\in \N$ we define the (partial) ``Hamming Ball Size'' function $\PS_n\colon [n] \times [0,1) \to [0,1)$ as follows
\begin{displaymath}
\PS_n(k,\lambda) = 2^{-n}\cdot\left(\sum_{i=0}^{k-1} \binom{n}{i} + \lambda \cdot \binom{n}{k} \right)\,.
\end{displaymath}
Note that this function  is a bijection and  we use $\PS^{-1}(\cdot)$ to denote its inverse. When $n$ is clear from the context, we will simply use $\PS(\cdot,\cdot)$ and  $\PS^{-1}(\cdot)$ instead.
\end{definition}

\remove{\begin{definition}
The function $\Phi\colon \R\to\R$ is the CDF of the normal distribution and is defined as follows:
$$\Phi(x) =\frac{1}{2}\left(1 + \erf\left(\frac{x}{\sqrt2}\right)\right).$$ 
\end{definition}}

\begin{lemma}\label{lem:mutok}
For $\mu \in [0,1]$ we have $\mu \geq \PS\left(\frac{n -\sqrt{-2\cdot\ln(\mu)\cdot n}}{2} +1,0\right)$. Also, if $(k,\lambda)=\PS^{-1}(\mu)$ then 
$$k\geq \frac{n -\sqrt{-2\cdot\ln(\mu)\cdot n}}{2}+1.$$
\end{lemma}
\begin{proof}
Let $k' = \frac{n -\sqrt{-2\cdot\ln(\mu)\cdot n}}{2}+1$. 
Now consider $n$ uniform random variables $X_1,\dots,X_n$ over $\set{0,1}$. We have
$$\PS(k',0) =\Pr[X_1+\dots+X_n \leq k'-1].$$ 
Then, by Hoefding inequality we have 
\begin{displaymath}
\Pr[X_1+\dots+X_n \leq k-1]
\leq \e^{-n\cdot{(1-\frac{2k-2}{n})^2}/{2}} = \mu
\end{displaymath}
Therefore we have,
$$\mu \geq \PS(k',0)$$ which proves the first part of the Lemma. Also the second part of the lemma immidiately follows because we have
$$\mu = \PS(k,\lambda) \geq \PS(k',0)$$ which implies $k\geq k'$.
\end{proof}

\begin{lemma}[Followed by the Central Limit Theorem~\cite{billingsley2008probability}] \label{lem:ham_cenlimit}
For all $\lambda\in[0,1)$ and $a\in \R$ we have
$$\lim_{n \to \infty}|\PS(n/2+a\cdot\sqrt{n}, \lambda)| = \Phi(2a)$$
where $\Phi$ is the CDF of the standard normall distribution.
\end{lemma}

\begin{lemma}[\cite{lovasz2003discrete}]\label{lem:lov}
If $1\leq k \leq \floor{n/2}$, then $\PS(k,0) < \binom{n}{k} \cdot {2^{2m-1 -n}}/{\binom{2m}{m}}$ where $m=\floor{\frac{n}{2}}$.
\end{lemma}
\begin{proof}
Case of $n=2m$ is Lemma 3.8.2 in \cite{lovasz2003discrete}. Case of $n=2m+1$ follows from the case for $n=2m$ and Pascal's equality.
\end{proof}
Using the above lemma, we can give a worst-case and a ``limit-case'' bound for the Hamming ball.

\begin{corollary} \label{lem:BallVolUp} If $1\leq k \leq \floor{n/2}$, then $\PS(k,0) < \binom{n}{k} \cdot \sqrt{\frac{n}{2^{2n+1}}}$. 
\end{corollary}
\begin{proof}
By Lemma \ref{fact:central_coef_lb} we know that
$\binom{2m}{m} \geq \frac{2^{2m-1}}{\sqrt{m}}$. Therefore, by Lemma \ref{lem:lov} we have
$\PS(k,0) < \binom{n}{k} \cdot \sqrt{\frac{n}{2^{2n+1}}}$.
\end{proof}
\begin{corollary}  \label{lem:BallVolUp_limit} 
For any $k\in\N$, $\lim_{n\to \infty} \frac{\PS(k,0)}{\sqrt{n}\cdot\binom{n}{k}\cdot 2^{-n}} \leq \sqrt{\frac{\pi}{8}}$.
\end{corollary}
\begin{proof}
Let $m=\floor{\frac{n}{2}}$. By Lemma \ref{lem:lov} we have
$$\frac{\PS(k,0)}{\sqrt{n}\cdot\binom{n}{k}\cdot 2^{-n}} \leq \frac{2^{2m-1}}{\sqrt{2m}\binom{2m}{m}}$$
which implies 
$\lim_{n\to \infty} \frac{\PS(k,0)}{\sqrt{n}\cdot\binom{n}{k}} \leq \lim_{m\to \infty} \frac{2^{2m-1}}{\sqrt{2m}\binom{2m}{m}} = \sqrt{\frac{\pi}{8}}$. Where the last equality follows from Lemma \ref{fact:central_coef_limit}.
\end{proof}

The following theorem, gives a general lower bound  for the adversarial risk of any classification problem for uniform distribution $U_n$ over the hypercube $\bits^n$, depending on the original error.
\begin{theorem}\label{thm:hq1}
Suppose $\problem=(\bits^n,\Y,U_n,\C,\hypoC,\HD)$ is a classification problem. For any $h \in \H, c\in \C $ and $r\in \N$, let $\mu=\Risk(h,c)>0$ be the original risk and $(k,\lambda) = \PS^{-1}\left(\mu\right)$ be a function of the original risk. Then, the error-region  adversarial risk under $r$-perturbation  is at least
\begin{displaymath}
\Risk\eDR_{r}(h,c) \geq  \PS(k+r,\lambda).
\end{displaymath}
\end{theorem}

Before proving Theorem  \ref{thm:hq1} we state and prove two corollaries. The proof of Theorem \ref{thm:hq1} appears in section \ref{sec:proofs_inherent}.

\begin{corollary}[Error-region risk for all $n$] \label{cor:hq3}
Suppose $\problem=(\bits^n,\Y,U_n,\C,\hypoC,\HD)$ is a classification problem. For any hypothesis $h,c$ with risk $\mu \in (0,\frac{1}{2}]$ in predicting a concept function $c$, we can increase the risk of $(h,c)$ from $\mu \in (0,\frac{1}{2}]$ to
$\mu'\in[\frac{1}{2},1]$ by changing at most 
$$r=\sqrt{\frac{-n \cdot \ln \mu}{2}}+\sqrt{\frac{-n \cdot \ln(1-\mu')}{2}}$$ 
bits in the input instances. Namely, by using the above $r$, we have $\Risk\eDR_r(h,c) \geq \mu'$. Also, to increase the error to $\frac{1}{2}$ we only need to change at most $r'=\sqrt{\frac{-n \cdot \ln(\mu) }{2}}$ bits.
\end{corollary}
\begin{proof}
 Let $(k,\lambda) = \PS^{-1}(\mu).$ By Theorem \ref{thm:hq1}, we know that 
 \begin{displaymath}
 \Risk\eDR_r(h,c) \geq \PS(k+r,\lambda)\,.
 \end{displaymath}
 By Lemma \ref{lem:mutok} we know that $k\geq \frac{n -\sqrt{-2\cdot\ln(\mu)\cdot n}}{2}$. 
Therefore we have
 \begin{align*}
 \Risk\eDR_r(h,c) &\geq \PS(k+r,\lambda) \\
 &\geq \PS\left(\frac{n +\sqrt{-2\cdot\ln(1-\mu')\cdot n}}{2},\lambda\right)\\ 
 &\geq 1 -\PS\left(\frac{n -\sqrt{-2\cdot\ln(1-\mu')\cdot n}}{2}+1,0\right)\\ 
 \text{(By Lemma \ref{lem:mutok})~~}
 &\geq \mu' \qedhere
 \end{align*}
Similarly, for the case of reaching error $\frac{1}{2}$ we have
 \begin{align*}
 \Risk\eDR_{r'}(h,c) &\geq \PS(k+r',\lambda)
 \geq \PS\left(\frac{n}{2},\lambda\right)
\geq  \frac{1}{2}. \qedhere
 \end{align*}
\end{proof}

\paragraph{Example.} Corollary \ref{cor:hq3} implies that for classification tasks over  $U_n$, by changing at most $3.04 \sqrt{n}$ 
number of bits in each example we can increase the error of an hypothesis from $1\%$ to $99\%$. Furthermore, for increasing the error just to $0.5$ we need half of the number of bits, which is $1.52 \sqrt n$. 

Also, the corollary bellow, gives a lower bound on the limit of adversarial risk when $n \To \infty$. This lower bound matches the bound we have in our computational experiments.
\begin{corollary}[Error-region risk for large $n$]\label{cor:hq3_limit}
Let  and $\problem=(\bits^n,\Y,U_n,\C,\hypoC,\HD)$ be a classification problem nd $\mu\in(0,1]$ and $\mu'\in(\mu,1]$. Then for any  $h \in \H, c\in \C $ such that $\Risk(h,c) \geq \mu$ we have  
$\Risk_r(h,c) \geq \mu'$ for 
$$r\approx \sqrt{n}\cdot\frac{\Phi^{-1}(\mu')-\Phi^{-1}(\mu)}{2} \text{ when } n\To \infty$$
 where $\Phi$ is the CDF of the standard normal distribution.
\end{corollary}

\begin{proof}
For simplicity suppose $\mu$ is exactly the risk (rather than a lower bound for it).
Let \\$(k,\lambda) = \PS^{-1}\left(\Risk(h,c)\right)$.
By Lemma \ref{lem:ham_cenlimit}, for $n \To \infty$ we have
$$\mu = |\PS(k,\lambda)| \approx \Phi\left(\frac{2k - n}{\sqrt{n}}\right).$$
Therefore, $k \approx n/2 + \Phi^{-1}(\mu)\cdot \sqrt{n}/2$. By Theorem \ref{thm:hq1} and another application of Lemma \ref{lem:ham_cenlimit}, 
\begin{align*}
\Risk_r(h,c) &\geq|\PS(k+r, 0)|\\
&\approx |\PS(n/2 + \frac{\Phi^{-1}(\mu)}{2}\cdot \sqrt{n} - \frac{\Phi^{-1}(\mu)}{2}\cdot \sqrt{ n} + \frac{\Phi^{-1}(\mu')}{2}\cdot\sqrt{n},0 )|\\
&= |\PS(n/2 + \frac{\Phi^{-1}(\mu')}{2}\cdot\sqrt{n},0)|\\
&\approx \Phi\left(\Phi^{-1}(\mu')\right)\\
&=\mu'
\end{align*}
\end{proof}

\paragraph{Example.} Corollary \ref{cor:hq3_limit} implies that for classification tasks over  $U_n$,  when $n$  is large enough, we can increase the error from $1\%$ to $99\%$ by changing at most $2.34\sqrt{n}$ bits, and we can  we can increase the error from $1\%$ to $50\%$ by changing at most $1.17\sqrt{n}$ bits in test instances.

The following theorem shows how to upper bound the adversarial \emph{robustness} using the original risk.
\begin{theorem}\label{thm:hq2}
Suppose $\problem=(\bits^n,\Y,U_n,\C,\hypoC,\HD)$ is a classification problem. For any $h \in \H$ and $c\in \C $, if $\mu = \Risk(h,c)$ and $(k,\lambda) = \PS^{-1}(\mu)$ depends on the original risk, then the error-region  robustness is at most
\begin{displaymath}
\Rob\eDR(h,c) \leq  \sum_{r=0}^{n-k+1} \left(1-\PS(k+r,\lambda)\right).
\end{displaymath}
\end{theorem}

Following, using Theorem \ref{thm:hq2}, we give an asymptotic lower bound for robustness before proving the Theorem. The proof of Theorem \ref{thm:hq2} appears in section \ref{sec:proofs_inherent}.
\begin{corollary} \label{cor:hq4}
Suppose $\problem=(\bits^n,\Y,U_n,\C,\hypoC,\HD)$ is a classification problem. For any hypothesis $h$ with risk $\mu \in (0,\frac{1}{2}]$, we can make $h$ to give always wrong answers by changing  $r={\sqrt{- n \cdot \ln \mu /2}} + \mu\cdot\sqrt{n/2}$ number of bits  on average. Namely, we have 
\begin{displaymath}
\Rob\eDR(h,c) \leq  \sqrt{\frac{- n \cdot \ln \mu}{2}} + \mu\cdot\sqrt{\frac{n}{2}}\,.
\end{displaymath}
And the following Corollary gives a lower bound on the robustness in limit.
\end{corollary}
\begin{proof}
Let $(k,\lambda)=\PS^{-1}(\mu)$. By Theorem~\ref{thm:hq2}, we have
\begin{displaymath}
\Rob\eDR(h,c) \leq \sum_{r=0}^{n-k+1} 1-\PS(k + r, \lambda)\leq \sum_{r=0}^{n-k+1} 1-\PS(k + r, 0).
\end{displaymath}
 On the other hand, by Lemma~\ref{lem:sum:binoms:i-times-binom} we know that $\sum_{i=1}^{n+1} \PS(i,0) = 1+ \frac{n}{2}.$ Therefore we have,
 \begin{displaymath}
 \Rob\eDR(h,c) \leq n-k+1 -\left(1 + \frac{n}{2}- \sum_{i=1}^{k-1} \PS(i,0)\right) = \frac{n}{2} -k+\sum_{i=0}^{k-1} \PS(i,0)\,.
 \end{displaymath}
Therefore we have,
 \begin{align*}
 \Rob\eDR(h,c) 
 &\leq \frac{n}{2} -k  +  \sum_{i=1}^{k-1} \PS(i,0)\addtocounter{equation}{1}\tag{\theequation} \label{ineq:0001}\\
 \text{(By Lemma \ref{lem:BallVolUp})~~}&\leq \frac{n}{2} -k + \sum_{i=1}^{k-1} \binom{n}{i}\cdot2^{-n}\cdot \sqrt{\frac{n}{2}}\\
&\leq \frac{n}{2} -k + \PS(k,0)\cdot\sqrt{\frac{n}{2}}\\
  &\leq  \frac{n}{2} -k + \mu\cdot \sqrt{\frac{n}{2}}. 
\end{align*}
On the other hand, by using Lemma~\ref{lem:mutok} we know that $k\geq \frac{n -\sqrt{-2\cdot\ln(\mu)\cdot n}}{2}+1$. Therefore, we have
\begin{displaymath}
\Rob\eDR(h,c)  \leq \sqrt{\frac{-n\cdot\ln \mu }{2}} + \mu\cdot\sqrt{\frac{n}{2}}\,.\qedhere
\end{displaymath}
\end{proof}

\begin{corollary} \label{cor:hq4_limit}
For any $\mu\in(0,1]$, classification problem $\problem=(\bits^n,\Y,U_n,\C,\hypoC,\HD)$, and any $h \in \H, c\in \C $ such that $\Risk(h,c) \geq \mu$, we have 
$$ \Rob\eDR(h,c) \leq  \frac{\Phi^{-1}(\mu)}{2}\cdot \sqrt{n} + \mu\cdot\sqrt{\frac{\pi\cdot n}{8}} \text{ ~~~when~~~ } n \To \infty,$$
where $\Phi$ is the CDF of the standard normall distribution.
\end{corollary}

\begin{proof}
Similar to inequality \ref{ineq:0001} in the proof of Corollary \ref{cor:hq4}, for large enough $n$ we have, 
 \begin{align*}
 \Rob\eDR(h,c) 
 &\leq \frac{n}{2} -k +  \sum_{i=1}^{k-1} \PS(i,0)\\
 \text{(By Corollary \ref{lem:BallVolUp_limit})~~}&\leq \frac{n}{2} -k + \sum_{i=1}^{k-1} \binom{n}{i}\cdot2^{-n}\cdot \sqrt{\frac{\pi\cdot n}{8}}\\
&\leq \frac{n}{2} -k + \PS(k,0)\cdot\sqrt{\frac{\pi\cdot n}{8}}\\
  &\leq  \frac{n}{2} -k +  \mu\cdot \sqrt{\frac{\pi \cdot n}{8}} 
\end{align*}
 Now, by Lemma $\ref{lem:ham_cenlimit}$, for large enough $n$ we have $k\approx  n/2 + \Phi^{-1}(\mu)\cdot \sqrt{n}/2$. Therefore we have
\begin{displaymath}
\Rob\eDR(h,c)  \leq \frac{\Phi^{-1}(\mu)}{2}\cdot\sqrt{n} + \mu\cdot\sqrt{\frac{\pi\cdot n}{8}}\,.\qedhere
\end{displaymath}
\end{proof}

\paragraph{Example.}  By changing $1.53 \sqrt{n}$ number of bits \emph{on average} we can increase the error of an hypothesis from $1\%$ to $100\%$. Also, if $n \To \infty$, by changing only $1.17 \sqrt{n}$ number of bits \emph{on average} we can increase the error from $1\%$ to $100\%$.

\iffollowingorders
\else
\subsection{Proving Theorems~\ref{thm:hq1} and~\ref{thm:hq2}}\label{sec:proofs_inherent}
We start by proving Theorem \ref{thm:hq1}. Before that we define several notations.
\begin{definition}[Volume, expansion, internal boundary and  external boundary] \label{def:VolEtAl-Disc}
Let $\cA \se \bits^n$.
\begin{itemize}
    \item The \emph{volume} of $\cA$ is simply  its probability $\vol(\cA)={|A|}/{2^n}$.
    \item The \emph{$r$-expansion} of $\cA$ is $\cA_r = \set{x\in \bits^n \mid \exists a\in A, \HD(a,x) \leq r}$
    \item The \emph{external boundary} of $\cA$  is defined as $\extB(\cA) = \cA_1 \sm \cA$.
    \item The \emph{internal boundary} of $\cA$ is  $\intB(\cA)=\set{a \mid a \in \cA, \exists x\in \bits^n \sm \cA, \HD(a,x)\leq 1}$.

\end{itemize} 
\end{definition}

The following isoperimetric inequality by Nigmatullin \citep{isoperiHQ} improves the famous vertex isoperimetric inequality of Harper \cite{harper1966optimal} in a way that is more suitable for us to use.

\begin{lemma}[\citep{isoperiHQ,harper1966optimal}]  \label{lem:bnd-ham} For any set $A \subset \bits^n$ where $(k, \lambda) = \PS^{-1}(\vol(A))$, we have
\begin{displaymath}
|\intB(A)| \geq \binom{n}{k-1} 
+ \lambda\cdot \left(\binom{n}{k} 
- \binom{n}{k-1} 
\right).
\end{displaymath}
\end{lemma}

The following simple lemma shows how to turn lower bounds for internal boundary into a lower bound for external boundary.
\begin{lemma}[External boundary vs. internal boundary]\label{lem:bnd-to-ext}
Suppose $b:[0,1] \to [0,1]$ is an internal boundary lower bound. Namely, For any given set $\cA\subseteq \bits^n$ we have,
\begin{displaymath}
\vol(\intB(\cA)) \geq b(\mu(\cA))\,.
\end{displaymath}
Then for any set $\cA \subseteq \bits^n$ we have,
\begin{displaymath}
\vol(\extB(\cA)) \geq \inf\set{b(\mu') \mid \mu'-b(\mu') \geq \vol(\cA)}\,.
\end{displaymath}
\end{lemma}

\begin{proof}
Suppose $\cA \se \bits^n$ is such that  $\vol(\cA) \geq \mu$, and let $\cA'=\cA \cup \extB(\cA)$. First we note that $\intB(\cA') \se \extB(\cA)$, because any point $a' \in \intB(\cA')$ cannot be also in $\cA$, as otherwise, $a'$ would not be an internal boundary point in $\cA'$.
Now, let $\mu' = \vol(\cA) =  \vol(\cA) + \vol(\extB(\cA))$. Because $b(\cdot)$ is an internal boundary lower bound, and because $\intB(\cA') \se \extB(\cA)$, therefore $b(\mu') \leq \vol(\extB(\cA))$. Thus, it holds that  $\mu'-b(\mu') \geq \mu' - \vol(\extB(\cA)) = \vol(\cA) \geq \mu$. By the definition of $\inf$, it holds that $ \inf\set{b(\mu') \mid \mu'-b(\mu') \geq \vol(\cA)} \leq b(\mu') $, which together with $b(\mu') \leq \vol(\extB(\cA))$ implies that $ \inf\set{b(\mu') \mid \mu'-b(\mu') \geq \vol(\cA)} \leq \vol(\extB(\cA))$.
\end{proof}

Now we  use  Lemma~\ref{lem:bnd-to-ext} to a derive a lower bound on the volume of exterior of a set.

\begin{lemma} \label{lem:ext-ham}
For any set $A \subset \bits^n$, if $(k, \lambda) = \PS^{-1}({|A|}/{2^n})$, then we have
\begin{displaymath}
\abs{\extB(A)} \geq \binom{n}{k} + \lambda\cdot \left(\binom{n}{k+1} - \binom{n}{k}\right).
\end{displaymath}
\end{lemma}

\begin{proof}[Proof of Lemma~\ref{lem:ext-ham}]
 Using Lemma~\ref{lem:bnd-ham} and Lemma~\ref{lem:bnd-to-ext} we have
\begin{displaymath}
|\extB(A)| \geq \inf\left\{ \binom{n}{k-1}  + \lambda\cdot \left(\binom{n}{k} - \binom{n}{k-1}\right) \mid \PS(k-1, \lambda)  \geq |A|\right\}.
\end{displaymath}
Since $\PS(\cdot)$ is a monotone function,
\begin{displaymath}
\abs{\extB(A)} \geq \binom{n}{k}  + \lambda\cdot\left( \binom{n}{k+1} - \binom{n}{k} \right).\qedhere
\end{displaymath}
\end{proof}

We now prove Theorem~\ref{thm:hq1}.
\begin{proof}[Proof of Theorem~\ref{thm:hq1}] Let set $\ERegion$ be the error region of hypothesis $h$ with respect to concept  $c$. Namely,
$\ERegion = \set {x\in \bits^n \mid h(x)\neq c(x)}$. Let $(k_\ERegion,\lambda_\ERegion) = \PS^{-1}({|\ERegion|}/{2^n})$. By Lemma~\ref{lem:ext-ham},
\begin{displaymath}
|\extB(\ERegion)| \geq \binom{n}{k_\ERegion}  + \lambda_\ERegion\cdot\left( \binom{n}{k_{\ERegion}+1} - \binom{n}{k_\ERegion} \right)\,.
\end{displaymath}
Now, by induction on $r$, we prove that $\vol(\ERegion_r) \geq \PS(k_\ERegion+r,\lambda_\ERegion)$. For $r=0$ the statement is trivially true. 
\remove{
we have,
\begin{align*}
\vol(\ERegion_1) &= \vol(\ERegion \cup \extB(\ERegion))\\
&= \vol(\ERegion) + \vol(\extB(\ERegion))\\
\text{(By Lemma~\ref{lem:ext-ham})}&\geq \PS(K_\ERegion,\lambda_\ERegion) + \binom{n}{k_\ERegion} + \lambda_\ERegion\cdot\left(\binom{n}{k_\ERegion+1} - \binom{n}{k_\ERegion} \right)\\
&= \PS(K_\ERegion+1, \lambda_\ERegion).
\end{align*}
}
Now for $r=i$, we assume $\vol(\ERegion_i) \geq \PS(k_\ERegion+i,\lambda_\ERegion)$. For $r=i+1$ we have,
\begin{align*}
\vol(\ERegion_{i+1}) &= \vol(\Ball_{i+1}(\ERegion))\\
&= \vol(\Ball_{i}(\ERegion)) + \vol(\extB(\Ball_{i}(\ERegion)))\\
\text{(By induction)~~} &\geq \PS(k_\ERegion+i, \lambda_\ERegion) + \vol(\extB(\Ball_{i}(\ERegion)))\\
\text{(By Lemma~\ref{lem:ext-ham})~~}&\geq \PS(k_\ERegion+i, \lambda_\ERegion) + \binom{n}{k_\ERegion+i} + \lambda_\ERegion\cdot\left( \binom{n}{k_\ERegion+i+1} - \binom{n}{k_\ERegion+i} \right)\\
&= \PS(k_\ERegion+i+1, \lambda_\ERegion).
\end{align*}
Therefore, the induction hypothesis is correct and we have $\vol(\ERegion_r) \geq \PS(k_\ERegion+r, \lambda_\ERegion)$ for all $r\in \N$. Note that $\vol(\ERegion_r)=\Risk\eDR_r(h,c)$, so we have
\begin{displaymath}
\Risk\eDR_r(h,c) \geq \PS(k_\ERegion+r, \lambda_\ERegion)\,.\qedhere
\end{displaymath}
\end{proof}

Now we state a useful lemma that connects risk to robustness.
\begin{lemma} \label{lem:newlem}
Suppose $\problem=(\bits^n,\Y,U_n,\C,\hypoC,\HD)$ is a classification problem. For any $h \in \H$ and $c\in \C $, if $(k,\lambda) = \PS^{-1}\left(\mu\right)$ for $\mu =\Risk(h,c)>0$, then we have,
\begin{displaymath}
\Rob\eDR(h,c) =  \sum_{r=0}^{n-k+1} \left(1-\Risk\eDR_r(h,c)\right).
\end{displaymath}
\end{lemma}
We now use the above lemma to prove Theorem \ref{thm:hq2}.
\begin{proof}
Let $\ERegion$ be the error region of $(h,c)$. Let $\Pr_{a \gets D}[\HD(a,\ERegion)=r] = p_r$. Then, it holds that
\begin{displaymath}
\Rob\eDR(h,c)= \sum_{r\geq 0} r \cdot p_r
=\sum_{r \geq 0} \sum_{i > r} p_i = \sum_{r \geq 0} (1-\vol(\ERegion_r)) = \sum_{r\geq 0} (1-\Risk\eDR_r(h,c))\,.
\end{displaymath}
\end{proof}
\begin{proof}[Proof of Theorem~\ref{thm:hq2}]
Based on Lemma~\ref{lem:newlem}, 
\begin{displaymath}
\Rob\eDR(h,c) = \sum_{r=0}^{n-k+1} (1-\Risk\eDR_r(h,c)).
\end{displaymath}
Now Theorem~\ref{thm:hq1} implies,
\begin{displaymath}
\Rob\eDR(h,c)\leq  \sum_{r=0}^{n-k+1} 1-\PS(k + r, \lambda).\qedhere
\end{displaymath}
\end{proof}

\fi

\section{Conclusions}
We discussed different definitions for adversarial perturbations. We saw that definitions that do not rely on the \emph{error region} do not guarantee misclassification for the perturbed (adversarial) instance in every case. Apart from the conceptual differences of the definitions, this was also shown through an extensive study of attacking monotone conjunctions under the uniform distribution \UUn (over $\bits^n$). In this study we were able to separate the adversarial robustness of hypotheses generated by popular algorithms, both in an asymptotic theoretical sense as well as in the average case by observing that the results of experiments also verify in practice the theoretical, worst-case, discrepancies on the robustness of the generated hypotheses. Subsequently, using the error region definition as the main definition for generating adversarial instances we were able to study and provide bounds for the adversarial risk and robustness of any classifier that is attacked when instances are drawn from the uniform distribution \UUn.


There are however many interesting questions for future work. 
One such example is theoretical and experimental investigation of popular algorithms in order to determine the adversarial risk and robustness of the generated solutions in various cases. This may involve specific distributions or classes of distributions and of course extend the investigation to concept classes beyond that of monotone conjunctions. Such investigations appear to be important because we can gain more insight on the geometry and the properties of the learned solutions and the extent to which such solutions are resistant to adversarial perturbations. For example, the solutions generated by \finds are consistently at least as resistant to adversarial perturbations (and for the largest part, significantly more resistant) as the solutions obtained by the \swapping.


Another question is of course to generalize the results that we have for the error region definition under the uniform distribution, to distributions beyond uniform. What kind of tools and techniques are required? In the case that such bounds contain the uniform distribution as a special case, how do such bounds compare to the bounds that we explored, given that our investigation is specific to the uniform distribution? Furthermore, our results of Section \ref{sec:uniform} are information theoretic. Are there equivalent computationally efficient attacks by adversaries that can achieve such bounds?


Getting along, our results, like all other current provable bounds in the literature for adversarial risk and robustness, only apply to specific distributions that do not cover the case of image distributions. These results, however, are first steps, and indicate similar phenomena (e.g., relation to isoperimetric inequalities). Thus, though challenging, these works motivate a deeper study of such inequalities for specific distributions. In addition, the work of Fawzi et al.~\citep{fawzi2018adversarial} has a discussion on how attacks on ``nice'' distributions can potentially imply attacks on real data. That would be the case assuming the existence of a generative model that transforms ``nicely''-distributed random latent vectors into natural data. Therefore, eventually an interesting line of work would be to investigate scenarios that also apply to the case of image classification and other very popular every-day classification tasks, highlighting further the importance of the investigations suggested in the previous paragraph.





\iffollowingorders
\bibliographystyle{plain}
\else
\bibliographystyle{alpha}
\fi

\newcommand{\etalchar}[1]{$^{#1}$}

\iffollowingorders
\else
\appendix
\section{Some Useful Facts}\label{sec:appendix:facts}

\iffollowingorders
\else
\begin{lemma}\label{lem:ratio-at-least-half}
Let $n\in\NN^*$. Then,
\begin{displaymath}
\sum_{i=0}^{\lfloor n/2\rfloor} \binom{n}{i} = 
\sum_{i=\lceil n/2\rceil}^n \binom{n}{i} \ge 2^{n-1}\,.
\end{displaymath}
\end{lemma}
\begin{proof}
If $n$ is even, then $n/2$ is an integer and moreover $\lfloor n/2\rfloor = n/2 = \lceil n/2\rceil$.
Therefore the sequence 
$\binom{n}{\lceil n/2\rceil}, \binom{n}{\lceil n/2\rceil + 1}, \ldots, \binom{n}{n}$
has $1 + n/2$ terms. The expansion $\binom{n}{0}, \binom{n}{1}, \ldots, \binom{n}{n}$ has $n+1$ terms
and moreover $\binom{n}{j} = \binom{n}{n-j}$ for $j\in\{0, 1, \ldots, n/2 - 1\}$. 
Let $A = \sum_{j=0}^{n/2 - 1} \binom{n}{j} = \sum_{k=n/2 + 1}^{n} \binom{n}{k}$.
Then, $A + \binom{n}{n/2} + A = 2^n \Rightarrow A = 2^{n-1} - \binom{n}{n/2}/2$.
But this implies that $\sum_{i=\lceil n/2\rceil}^n \binom{n}{i} = \binom{n}{n/2} + A = 2^{n-1} + \binom{n}{n/2}/2 > 2^{n-1}$.

If $n$ is odd, then the sequence 
$\binom{n}{\lceil n/2\rceil}, \binom{n}{\lceil n/2\rceil + 1}, \ldots, \binom{n}{n}$
has $1 + n - \lceil n/2\rceil = 1 + \lfloor n/2\rfloor = (1+n)/2$ terms.
Further, again by symmetry we have 
$B = \sum_{j=0}^{\lfloor n/2\rfloor} \binom{n}{j} = \sum_{k=\lceil n/2\rceil}^{n} \binom{n}{k}$.
Then, $B + B = 2^n \Rightarrow B = 2^{n-1}$.
\end{proof}
\fi

%
%
\iffollowingorders
\else
\begin{lemma}\label{fact:central_coef_lb} 
For any $m\in\N$ we have $ \binom{2m}{m}  \geq \frac{2^{2m-1}}{\sqrt{m}}$.
\end{lemma}
%
%
\begin{lemma}\label{fact:central_coef_limit}
We have $\lim_{m\to\infty} \frac{2^{2m}}{\binom{2m}{m}\cdot \sqrt{m}} = \sqrt{{\pi}}.$
\end{lemma}
\fi

%
%
\iffollowingorders
\else
\begin{lemma}\label{lem:sum:binoms:i-times-binom}
Let $k\in\NN$. Then, 
$\sum_{i=0}^k i\cdot\binom{k}{i} = k\cdot2^{k-1}$.
\end{lemma}
\fi

\iffollowingorders
\else
\begin{lemma}\label{lem:important:lower-bound}
Let $1 \le u \le w$.
Let $1\le \zeta \le u$ and $1\le \xi \le w$.
Then, 
\begin{displaymath}
\frac{u}{8}\cdot 2^{u+w} 
\le 
\sum_{\zeta=1}^u \binom{u}{\zeta} \sum_{\xi=\zeta}^w \binom{w}{\xi}\min\{\zeta, \xi\} 
\end{displaymath}
\end{lemma}
\begin{proof}
%
We have, 
\begin{align*}
\sum_{\zeta=1}^u \binom{u}{\zeta} \sum_{\xi=\zeta}^w \binom{w}{\xi}\min\{\zeta, \xi\}~
  &\ge \sum_{\zeta=1}^{\lceil u/2\rceil} \zeta\binom{u}{\zeta} \sum_{\xi=\lceil w/2\rceil}^{w} \binom{w}{\xi} \nonumber\\
  &\ge ~2^{w-1}\sum_{\zeta=1}^{\lceil u/2\rceil} u\binom{u-1}{\zeta-1} 
  & \mbox{(Lemma \ref{lem:ratio-at-least-half} 
  })\nonumber\\
  &= ~\frac{u}{2}\cdot 2^{w}\cdot \sum_{\zeta=0}^{\lceil u/2\rceil-1} \binom{u-1}{\zeta} \nonumber\\
  &= ~\frac{u}{2}\cdot 2^{w}\cdot \sum_{\zeta=0}^{\lfloor (u-1)/2\rfloor} \binom{u-1}{\zeta} \nonumber 
\end{align*}
We now use Lemma \ref{lem:ratio-at-least-half} again and obtain the statement.
\end{proof}
\fi

\fi

\end{document}